\documentclass[journal]{IEEEtran}
\usepackage{amsmath,cite}
\usepackage{amssymb}
\usepackage{epsfig}
\usepackage[ruled,vlined,linesnumbered]{algorithm2e}
\usepackage{algorithmic}
\usepackage{subfigure}
\usepackage{booktabs,makecell}
\usepackage{multirow}
\usepackage{threeparttable}
\usepackage{color}
\usepackage{stfloats}
\usepackage{lipsum}
\usepackage{cuted}
\usepackage{url}
\usepackage{footnote}

\ifCLASSINFOpdf
\else
\fi
\begin{document}
%
\title{Error Loss Networks}
%
%
%

\author{Badong~Chen,~\IEEEmembership{Senior Member,~IEEE,}
        Yunfei~Zheng,~\IEEEmembership{}
        and Pengju~Ren,~\IEEEmembership{Member, IEEE}
\thanks{Badong Chen (corresponding author), Yunfei Zheng, and Pengju Ren are with the
Institute of Artificial Intelligence and Robotics, Xi'an Jiaotong University,
Xi'an 710049, China (e-mail: chenbd@mail.xjtu.edu.cn; zhengyf@stu.xjtu.edu.cn; pengjuren@gmail.com).}
\thanks{This work was supported by the National Natural Science Foundation of China under Grant Numbers (U21A20485, 61976175). }
}

\maketitle

\begin{abstract}
A novel model called error loss network (ELN) is proposed to build an error loss function for supervised learning. The ELN is in structure similar to a radial basis function (RBF) neural network, but its input is an error sample and output is a loss corresponding to that error sample. That means the nonlinear input-output mapper of ELN creates an error loss function. The proposed ELN provides a unified model for a large class of error loss functions, which includes some information theoretic learning (ITL) loss functions as special cases. The activation function, weight parameters and network size of the ELN can be predetermined or learned from the error samples. On this basis, we propose a new machine
learning paradigm where the learning process is divided into two
stages: first, learning a loss function using an ELN; second, using
the learned loss function to continue to perform the learning. Experimental results are presented to demonstrate the desirable performance of the new method.


\end{abstract}

%
\IEEEpeerreviewmaketitle

%
%
%
\section{Introduction}
\begin{figure}[htbp]
  \centering
  \includegraphics[width=7.5cm]{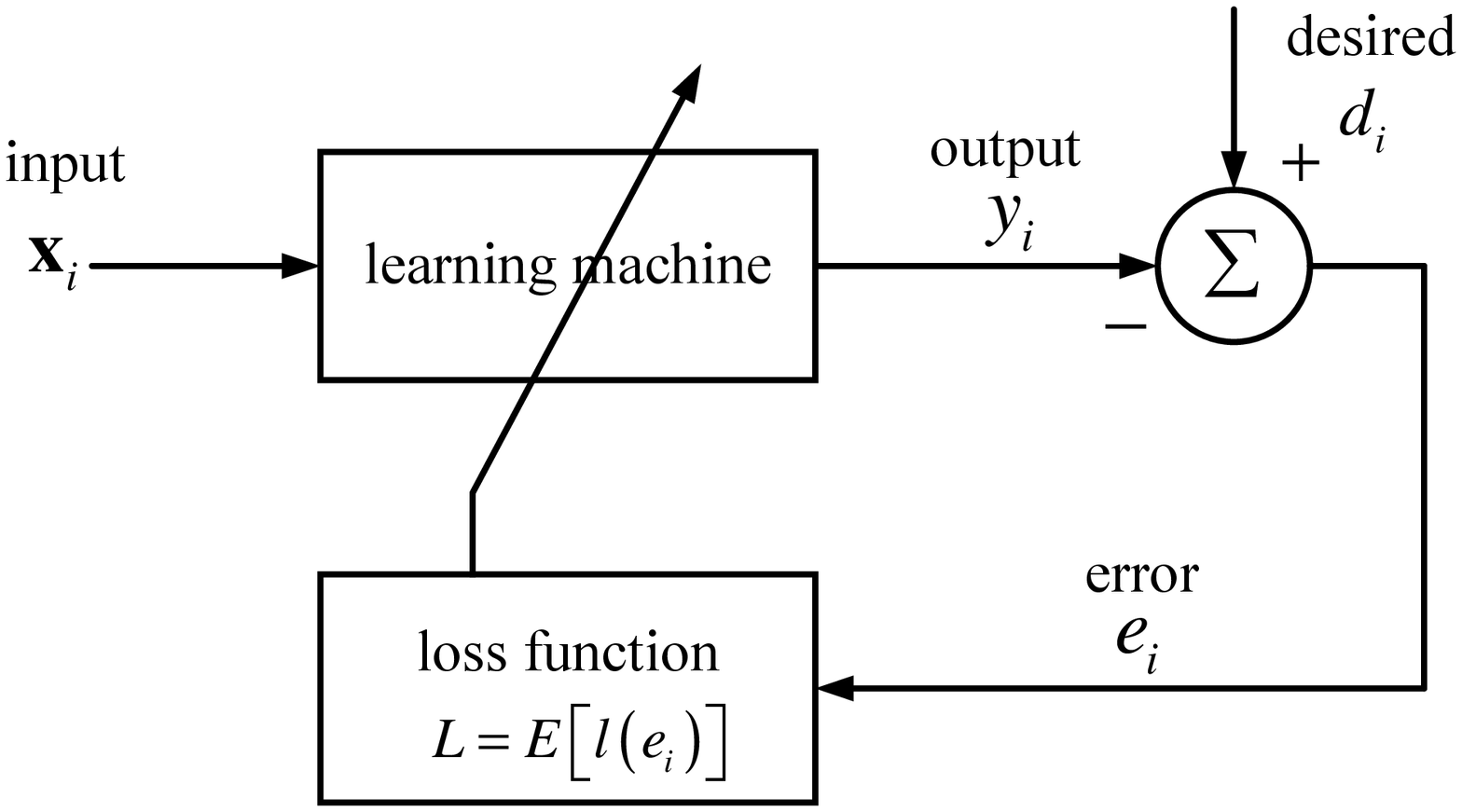}
  \caption{Schematic of supervised learning.}
  \label{fig1}
\end{figure}

Machine learning aims to build a model based on training samples to predict the output of new samples. For supervised learning (see Fig.~\ref{fig1}), each training sample consists of a pair of data called respectively the input (typically a vector) and the desired output (also called the supervisory signal). Given training samples $\left\{ {\left( {{\mathbf{x}_1},{d_1}} \right), \cdots ,\left( {{\mathbf{x}_N},{d_N}} \right)} \right\}$ with ${\mathbf{x}_i}$ being the input vector and ${d_i}$ the desired output, a supervised learning algorithm usually seeks a function (i.e. the input-output mapper of the learning machine) $f:\mathcal{X} \to \mathcal{Y}$ where $\mathcal{X}$ denotes the input space and $\mathcal{Y}$ stands for the output space, such that an empirical loss (risk) is minimized, that is
\begin{equation}\label{E1}
{f^*} = \mathop {\arg \min }\limits_{f \in \mathcal{F}} \frac{1}{N}\sum\limits_{i = 1}^N {l\left( {{d_i},{y_i}} \right)},
\end{equation}
where $\mathcal{F}$ is some space of possible functions (usually called the hypothesis space), ${y_i} = f\left( {{\mathbf{x}_i}} \right)$ is the model (function) output, and $l:\mathcal{Y} \times \mathcal{Y} \to \mathbb{R}$ is a certain loss function that measures how well a function fits the training samples (In this work, the loss function is not limited to be nonnegative). In many cases, the loss function depends on the error sample ${e_i} = {d_i} - {y_i}$, namely the difference between the desired and the model output. In such conditions, we call it an error loss, denoted by $l\left( {{e_i}} \right)$. Some regularization terms are often incorporated into the loss function to prevent overfitting. However, in this study our focus is mainly on the error loss term.

How to choose or design a proper error loss function is a key problem in supervised learning. The squared error loss (i.e. the mean square error (MSE)), in which , $l\left( e \right) = {e^2}$ is one of the most widely used loss functions especially for regression problem because of its mathematical convenience and computational simplicity. However, MSE cannot deal with non-Gaussian noise well, and particularly is sensitive to heavy-tailed outliers. Many alternatives such as mean absolute error (MAE) \cite{MAE}, mean p-power error (MPE) \cite{MPE1, MPE2, MPE3}, Huber's loss \cite{Huber1, Huber2} and logarithmic loss \cite{Logarithmic1, Logarithmic2, Logarithmic3}, can thus be used to improve the robustness to outliers. Over the past decade, some quantities (such as entropy) related to information theory have been successfully used as robust loss functions in machine learning, and such learning methods are called information theoretic learning (ITL) \cite{ITL}. The minimum error entropy (MEE) \cite{MEE-Supervised, MEE-ADALINE} and maximum correntropy criterion (MCC) \cite{P-MCC} are two typical learning criteria in ITL. By Parzen window approach, the ITL loss functions can be estimated directly from the samples, and with a Gaussian kernel these loss functions can be viewed as a similarity measure in kernel space \cite{ITL}. Compared with other robust error loss functions, such as MAE and Huber's loss, the ITL loss functions often show much better robustness to complex non-Gaussian noises e.g., heavy-tailed, multimode distributed, discrete-valued, etc.). Theoretical analysis of the robustness of the ITL can be found in \cite{MEE-Robustness, MCC-Robustness1, MCC-Robustness2}.

In recent years, several important variants or extended versions of the ITL learning criteria (loss functions) have been proposed to further enhance the robustness or improve the computational efficiency. Examples are quantized minimum error entropy (QMEE) \cite{QMEE}, generalized maximum correntropy criterion (GMCC) \cite{GMCC}, maximum mixture correntropy criterion (MMCC) \cite{MMCC1, MMCC2}, kernel risk-sensitive loss (KRSL) \cite{KRSL}, kernel mean p-power error (KMPE) \cite{KMPE}, maximum correntropy criterion with variable center (MCC-VC) \cite{MCC-VC}, maximum multi-kernel correntropy criterion (MMKCC) \cite{MMKCC}, and so on. These ITL loss functions have been successfully applied to robust regression, adaptive filtering, principal component analysis (PCA), point set registration, Granger causality analysis, etc. \cite{MCC-Regression, KMC, RIF-MCC, MCUF, KRMC, RHAF-MCC, SMCC, MCKF, MCC-PCA, MCC-2D, KMEE, QMEE-Granger, MCC-Granger}. How to integrate these loss functions into a unified model is of great significance, but also a great challenge.

In the present paper, we propose a novel model called error loss network (ELN) to build an error loss function for supervised learning. The ELN is in structure similar to a radial basis function (RBF) neural network, but its input is an error sample and output is a loss corresponding to that error sample. That means the nonlinear input-output mapper of ELN creates an error loss function. The proposed ELN provides a unified model for a large class of error loss functions, including the previously mentioned ITL loss functions as special cases. The activation function, weight parameters and network size of the ELN can be predetermined or learned from the error samples. On this basis, we propose a new machine
learning paradigm where the learning process is divided into two
stages: first, learning a loss function using an ELN; second, using
the learned loss function to continue to perform the learning.
Moreover, we propose a probability density function (PDF) matching approach to train the ELN and a fixed-point iterative algorithm to train the learning machine with a linear-in-parameter (LIP) model.

The rest of the paper is organized as follows. In section II, we give the basic concept of ELN and discuss its relationship with ITL. In section III, we propose an overall supervised learning strategy and discuss how to train the ELN and the learning machine. Section IV presents the experimental results and finally, section V gives the concluding remarks.

\section{Error Loss Networks} \label{Sec2}
\subsection{Basic Concept} \label{Sec2-1}
\begin{figure}[htbp]
  \centering
  \includegraphics[width=7.5cm]{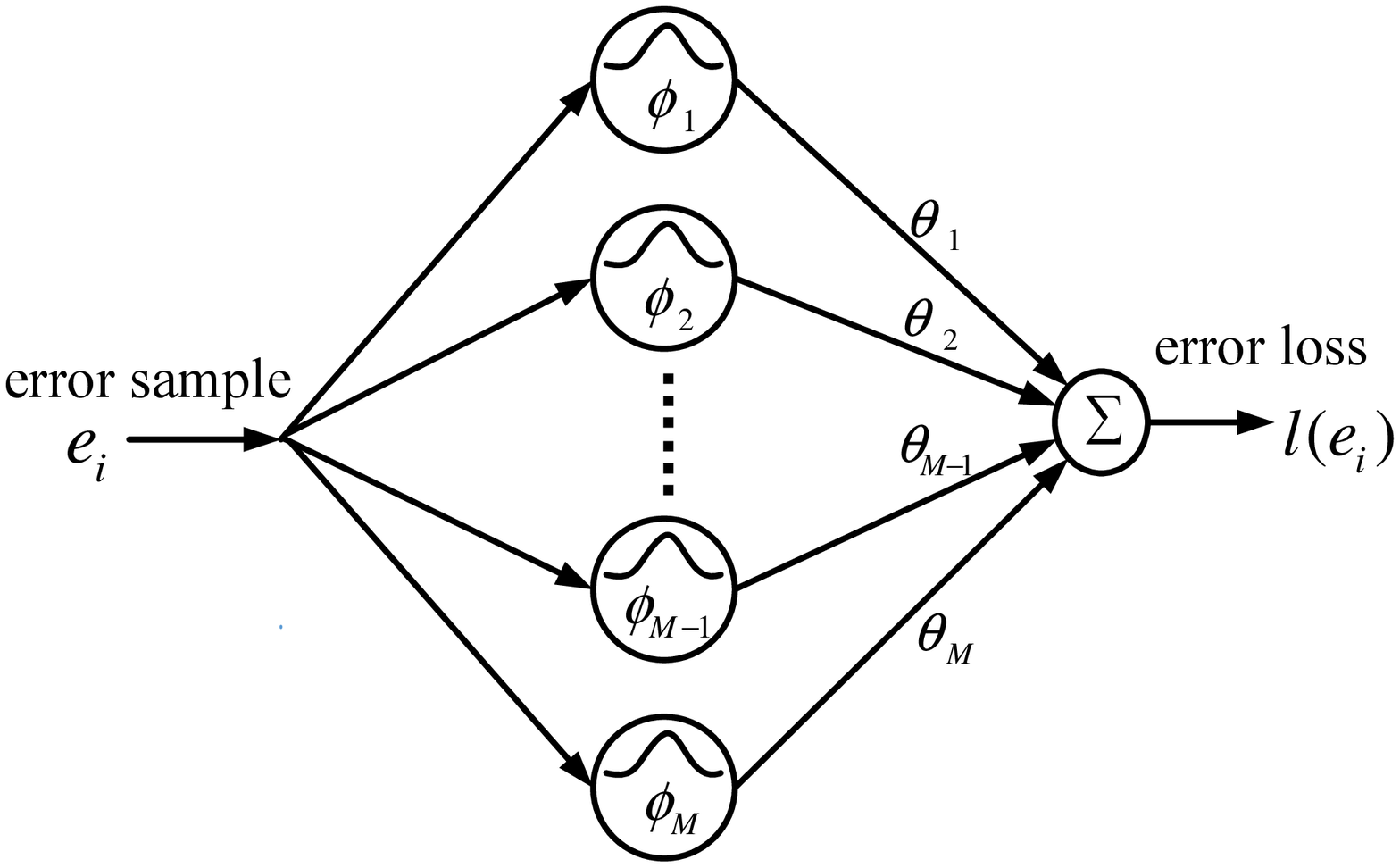}
  \caption{Schematic of error loss network.}
  \label{fig2}
\end{figure}

The error loss is generally a nonlinear function of the error sample (the difference between the desired and the model output), which has very important influence on the performance of supervised learning. In most applications, such functions are predetermined by the algorithm designers based on experience or some prior knowledge. For example, one often chooses $l\left( e \right) = {e^2}$ for ordinary regression or $l\left( e \right) = \left| e \right|$ for robust regression. In this work, we propose to build an error loss function in a manner similar to an RBF neural network, which can be trained from the data (not predetermined). Specifically, we propose an RBF like model called ELN, with error sample as the input and corresponding output as the loss (see Fig.~\ref{fig2}). The nonlinear input-output mapper of ELN creates an error loss function $l(.)$.

Let $\left\{ {{\phi _1}, \cdots ,{\phi _M}} \right\}$ be $M$ radial basis functions (whose values depend only on the distances between the inputs and some fixed points) and $\boldsymbol{\theta}={\left[ {{\theta _1}, \cdots ,{\theta _M}} \right]^T}$ be the output weight vector. The output (error loss) of the ELN is a linear combination of the radial basis functions, given by
\begin{equation}\label{E2}
l\left( {{e_i}} \right) = \sum\limits_{j = 1}^M {{\theta _j}{\phi _j}\left( {{e_i}} \right)}.
\end{equation}
Given $N$ error samples $\left\{ {{e_1}, \cdots ,{e_N}} \right\}$, the empirical loss based on the ELN can thus be computed by
\begin{equation}\label{E3}
L = \frac{1}{N}\sum\limits_{i = 1}^N {l\left( {{e_i}} \right)}  = \frac{1}{N}\sum\limits_{i = 1}^N {\sum\limits_{j = 1}^M {{\theta _j}{\phi _j}\left( {{e_i}} \right)} }.
\end{equation}

The ELN model can be extended to other neural networks (e.g. Multilayer Perceptron), but this paper is concerned only with the RBF type network. It is worth noting that if the radial basis functions (e.g., Gaussian function) satisfy: i) $\forall e \in \mathbb{R}, {\phi _j}\left( e \right) \le {b_j},{b_j} > 0$; ii) $\mathop {\lim }\limits_{\left| e \right| \to \infty } \frac{\partial }{{\partial e}}{\phi _j}\left( e \right) = 0$, we have
\begin{equation}\label{E4}
\left\{ \begin{array}{l}
\left| {l(e)} \right| \le \sum\limits_{j = 1}^M {\left| {{\theta _j}} \right|} {b_j}\\
\mathop {\lim }\limits_{\left| e \right| \to \infty } \frac{\partial }{{\partial e}}l\left( e \right) = 0 .
\end{array} \right.
\end{equation}
In this case, the error loss function $l(e)$ is always bounded and its derivative will approach zero as $\left| e \right| \to \infty $. That means the influence of an error with very large value on the loss function is very limited, because the loss at large error is bounded and the gradient of the loss function at large error is also very small. Such loss function will be robust to outliers that usually cause large errors. Therefore, with an ELN one can easily obtain an error loss function robust to outliers by choosing the radial basis functions satisfying the previous two conditions. This is a great advantage of the proposed ELN.

\subsection{Relation to ITL}
The ELN has close relationship with the celebrated ITL, initiated in the late 90s. Basically, ITL uses descriptors from information theory (e.g., entropy) estimated directly from the data to create the loss functions for machine learning. One of the most popular ITL learning criteria is the MEE, which adopts the error's entropy as the loss for supervised learning (to say, the learning machine is trained such that the error entropy is minimized). With Renyi's quadratic entropy, the empirical loss of MEE can be computed by \cite{ITL, MEE-Supervised, MEE-ADALINE}
\begin{align}\label{E5}
L &=  - \int\nolimits_{ - \infty }^{ + \infty } {{{\left( {\hat p(e)} \right)}^2}d} e \nonumber\\
   &=  - \int\nolimits_{ - \infty }^{ + \infty } {{{\left( {\frac{1}{N}\sum\limits_{i = 1}^N {{G_\sigma }\left( {e - {e_i}} \right)} } \right)}^2}de} \nonumber\\
 &= -\frac{{  1}}{N^2}\sum\limits_{i = 1}^N {\sum\limits_{j = 1}^N {{G_{\sqrt 2 \sigma }}\left( {{e_i} - {e_j}} \right)} },
\end{align}
where $\hat p(e){\rm{ = }}\frac{1}{N}\sum_{i = 1}^N {{G_\sigma }\left( {e - {e_i}} \right)} $ is the estimated PDF (by Parzen window approach \cite{Parzen1962}) of the error based on the $N$ samples $\left\{ {{e_1}, \cdots ,{e_N}} \right\}$, and ${G_\sigma }(e)$ is a Gaussian kernel controled by kernel width $\sigma$ and can be expressed by
\begin{equation} \label{GaussFun}
	{G_\sigma }(e) = \frac{1}{\sqrt{2 \pi}\sigma} \exp \left( { - \frac{{{e^2}}}{{2{\sigma ^2}}}} \right).
\end{equation}	
The above empirical loss can be rewritten as
\begin{align}\label{E6}
L & = \frac{1}{N}\sum\limits_{i = 1}^N {\left( {\sum\limits_{j = 1}^N {\frac{{ - 1}}{{N}}{G_{\sqrt 2 \sigma }}\left( {{e_i} - {e_j}} \right)} } \right)} \nonumber\\
& = \frac{1}{N}\sum\limits_{i = 1}^N {\left( {\sum\limits_{j = 1}^N {{\theta _j}{\phi _j}\left( {{e_i}} \right)} } \right)},
\end{align}
where ${\theta _j} = -\frac{{1}}{{N }}$, ${\phi _j}\left( {{e_i}} \right) = {G_{\sqrt 2 \sigma }}\left( {{e_i} - {e_j}} \right)$. Thus, the error loss function $l(e)$ of MEE can be created by an ELN model with $N$ hidden nodes ${\phi _j}\left( e \right) = {G_{\sqrt 2 \sigma }}\left( {e - {e_j}} \right), j = 1, \cdots ,N$, and output weight vector $\boldsymbol{\theta}  = {\left[ { -\frac{{ 1}}{{N}}, \cdots ,- \frac{{ 1}}{{N }}} \right]^T}$.

The network size of the ELN for MEE is equal to the sample number, which is very large for large scale data sets. To reduce the computational complexity of MEE, the quantized MEE (QMEE) was proposed in \cite{QMEE}. Similar to the MEE, the QMEE loss can also be viewed as an ELN, but with network size $M \leq N$ in general. The basic idea behind QMEE is actually to merge several hidden nodes into one if their centers are very close.
\begin{table*}[htbp]
	\renewcommand\arraystretch{1.3}
	\newcommand{\tabincell}[2]{\begin{tabular}{@{}#1@{}}#2\end{tabular}}
	\caption{Error loss networks of some ITL loss functions}
	\begin{center}
		\begin{tabular}{cccc}
			\toprule
			Loss & $M$   & ${\phi _i}(i = 1, \cdots ,M)$ & ${\theta _i}(i = 1, \cdots ,M)$   \\
			\midrule
			MCC \cite{P-MCC} & $M=1$  & ${\phi _1}(e) = {G_\sigma }(e)$ & ${\theta _1} =  - 1$   \\
			GMCC \cite{GMCC} & $M=1$   & ${\phi _1}(e) = {G_{\alpha ,\beta }}(e)$  & ${\theta _1} =  - 1$   \\
			KRSL \cite{KRSL} & $M=1$   & ${\phi _1}(e) = \exp \left( {\lambda \left( {1 - {G_\sigma }(e)} \right)} \right)$ & ${\theta _1} = \frac{1}{\lambda }$   \\
			KMPE \cite{KMPE} & $M=1$   & ${\phi _1}(e) = {\left( {1 - {G_\sigma }(e)} \right)^{{p \mathord{\left/
							{\vphantom {p 2}} \right.
							\kern-\nulldelimiterspace} 2}}}$ & ${\theta _1} = 1$   \\
			MCC-VC \cite{MCC-VC}    & $M=1$   & ${\phi _1}(e) = {G_\sigma }(e - c)$ & ${\theta _1} = -1 $   \\
			MMCC (Type 1) \cite{MMCC1}    & $M=2$   & $
({\phi _1}(e), {\phi _2}(e)) = ({G_{{\sigma _1}}}(e),
			 {G_{{\sigma _2}}}(e))$ & $
			({\theta _1}, {\theta _2}) =  (- \alpha, \alpha  - 1)
			$   \\
			MMCC (Type 2) \cite{MMCC2}     & $M=2$  & $
			({\phi _1}(e), {\phi _2}(e)) = ({G_{{\sigma}}}(e), {L_{{\sigma}}}(e))$ & $
			({\theta _1}, {\theta _2}) =  (- \alpha, \alpha  - 1)
			$   \\
			MEE \cite{MEE-Supervised}  & $M=N$   & ${\phi _i}(e) = {G_\sigma }(e - {e_i})$ & ${\theta _i} = \frac{ - 1}{N}$   \\
			QMEE \cite{QMEE}  & $1\leq M\leq N $   & ${\phi _i}(e) = {G_\sigma }(e - {c_i})$ & ${\theta _i} = \frac{{ - {M_i}}}{{N}}$   \\
			RMEE \cite{RMEEL}  & $M=3 $   & $({\phi _1}(e), {\phi _2}(e), {\phi _3}(e)) \!=\! ({G_\sigma }(e \!-\! 0),{G_\sigma }(e \!+\!1), {G_\sigma }(e \!-\! 1)) $  & $({\theta _1}, {\theta _2},{\theta _3}) \!=\! (\frac{{ - {M_1}}}{{N}},\frac{{ - {M_2}}}{{N}}, \frac{{ - {M_3}}}{{N}})$  \\
			MMKCC \cite{MMKCC}  & $M\geq2$   & ${\phi _i}(e) = {G_{{\sigma _i}}}(e - {c_i})$ & ${\theta _i} = -\lambda_i $   \\
			\bottomrule
		\end{tabular}
        \begin{tablenotes}
        \footnotesize
        \item[1] \hspace{0.2cm} The symbol $L_{{\sigma}}$ denotes an Laplacian kernel controlled by the kernel size $\sigma$.
      \end{tablenotes}
		\label{Tab1}
	\end{center}
\end{table*}

Another popular learning criterion in ITL is the MCC, which is computationally much simpler than the MEE criterion (but the performance of MEE is usually better in the case of complex noise). The empirical loss under MCC is \cite{P-MCC}
\begin{equation}\label{E7}
L =  - \frac{1}{N}\sum\limits_{i = 1}^N {{G_\sigma }\left( {{e_i}} \right)}.
\end{equation}
There is a minus sign in the above formula because the loss is minimized. One can see that the error loss function of MCC is $l(e) =  - {G_\sigma }(e)$ (sometimes $l(e) = 1 - {G_\sigma }(e)$ is used to ensure the non-negativity of the loss function). Thus, the MCC loss function can be viewed as a special ELN with only one hidden node ${\phi _1}(e) = {G_\sigma }(e)$ and corresponding output weight ${\theta _1} =  - 1$.

Recently, in order to further improve the learning performance, the MCC loss function has been extended to non-Gaussian kernel functions (e.g., GMCC \cite{GMCC}, KRSL \cite{KRSL}, KMPE \cite{KMPE}) or multi-kernel functions (e.g., MMCC \cite{MMCC1, MMCC2}, MMKCC \cite{MMKCC}). In fact, all these loss functions can be integrated into a unified model, the ELN. The network size, radial basis functions and output weights of the ELNs for several ITL loss functions are summarized in Table~\ref{Tab1}.

\section{Supervised Learning with ELNs}
\subsection{Overall Learning Strategy}
\begin{figure}[htbp]
  \centering
  \includegraphics[width=7.5cm]{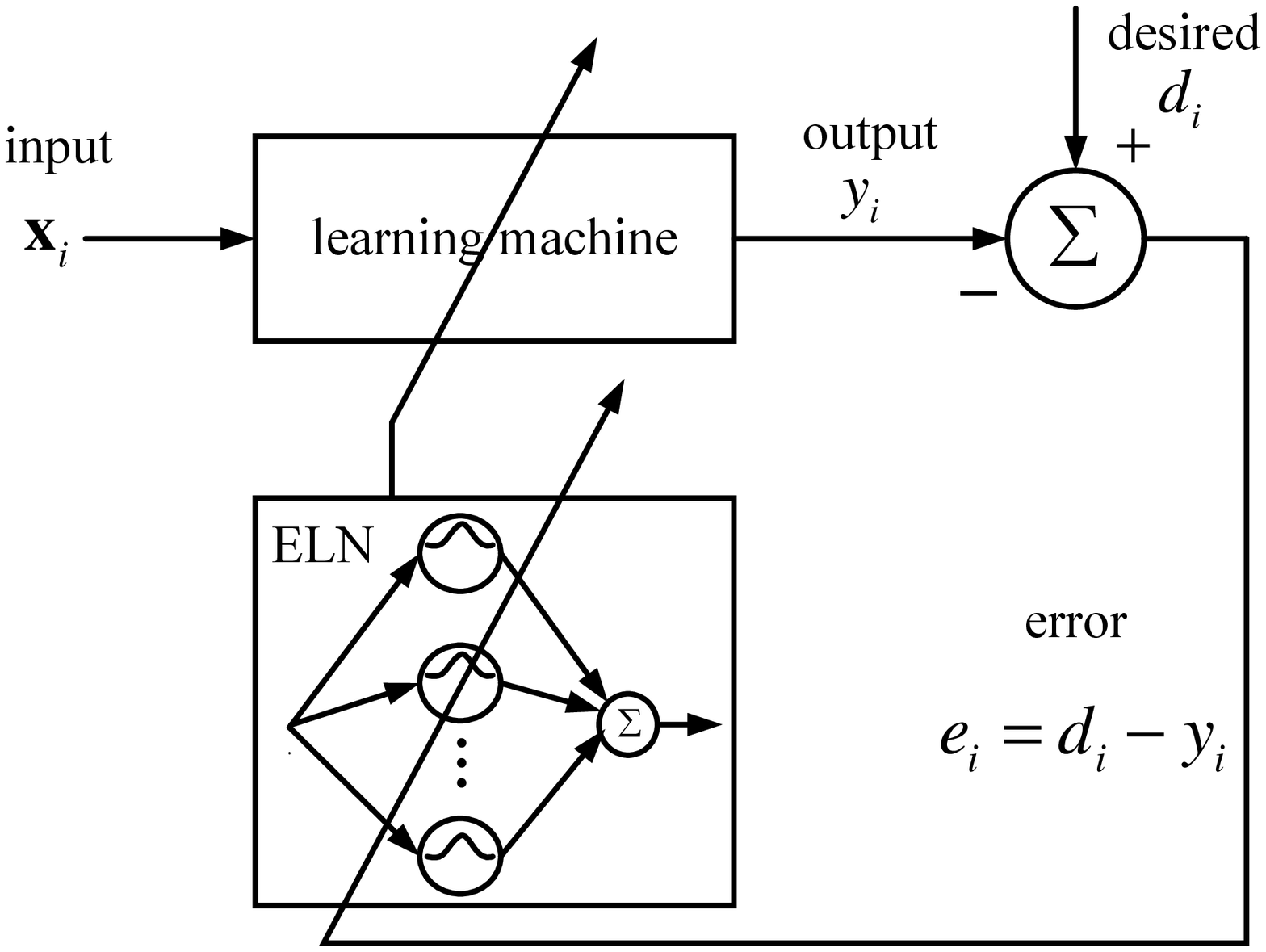}
  \caption{Supervised learning with ELN.}
  \label{fig3}
\end{figure}

When applying the ELN to supervised learning, before training the learning machine one must have an ELN available. To this end, one can predetermine an ELN (e.g., assign an existing ITL loss function) or use the training data to train an ELN. In this study, we mainly consider the second case, i.e., training an ELN to obtain a data-driven loss function for supervised learning. In this case, the whole learning process consists of two sub-learning processes, i.e., training the ELN and training the learning machine (see Fig.~\ref{fig3} for the general schematic).

\begin{figure}[htbp]
  \centering
  \includegraphics[width=7.5cm]{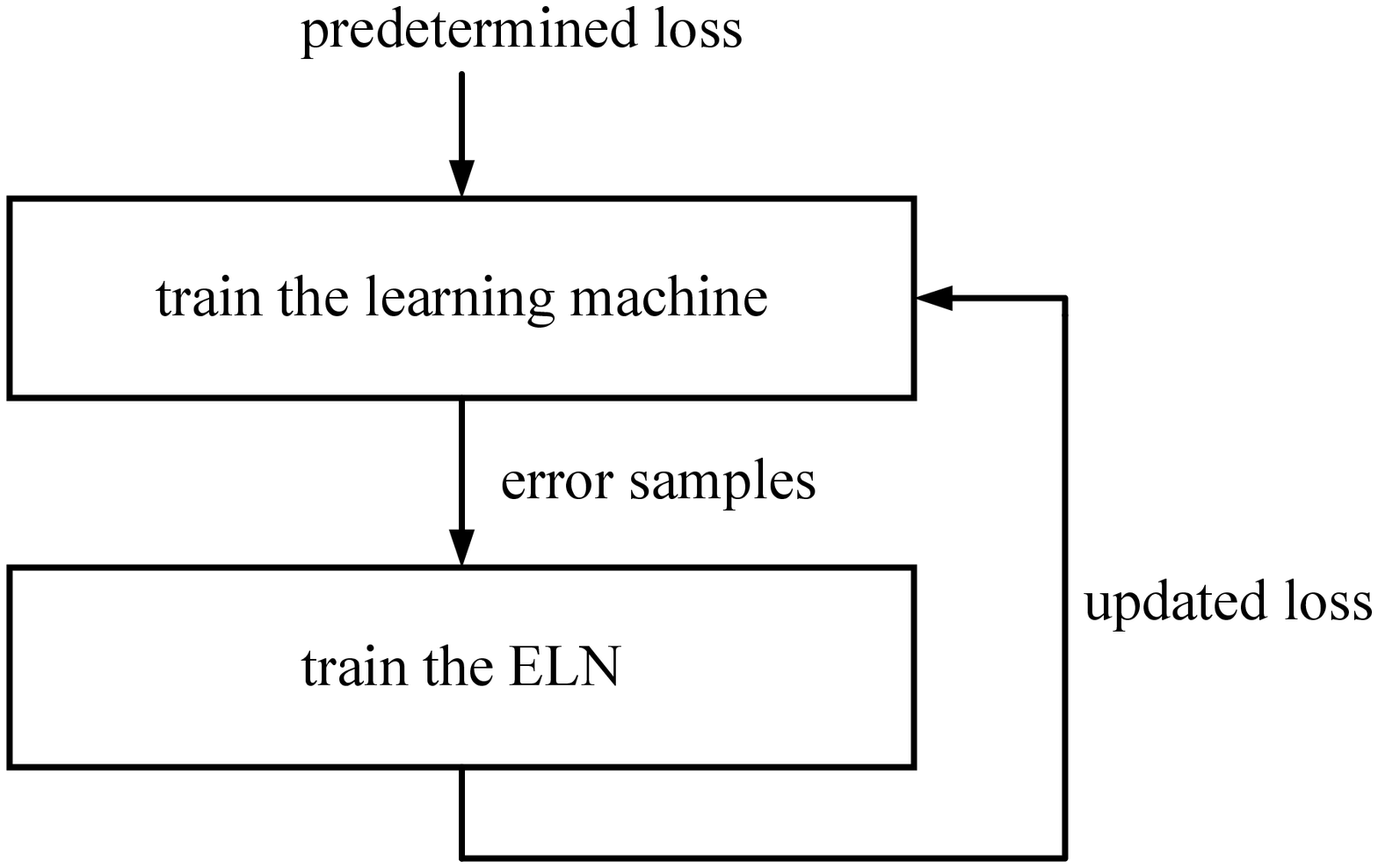}
  \caption{Schematic of the two-stage learning with ELN.}
  \label{fig4}
\end{figure}

The two sub-learning processes in Fig.~\ref{fig3} are interdependent, that is, training the ELN depends on the error samples from the trained learning machine; training the learning machine depends on the loss function from the trained ELN. How to combine the two sub-learning processes to form a complete learning process is a key problem. To solve this problem, we first assume that the initial error samples over the training data set have been obtained with a predetermined loss function, and then perform the following two sub-learning processes: first, using the available error samples to train an ELN and obtain an updated error loss function; second, using
the updated loss function to continually train the learning machine.
The two processes can form a cycle, whose schematic is shown in Fig.~\ref{fig4}.

\subsection{Training the ELN}
The proposed ELN model consists of two key parts. The first is the weight vector $\boldsymbol{\theta}={\left[ {{\theta _1}, \cdots ,{\theta _M}} \right]^T}$ and the second is $M$ radial basis functions $\left\{ {{\phi _1}, \cdots ,{\phi _M}} \right\}$.
In this subsection, we discuss how to use the error samples to train the ELN model.

\subsubsection{Determine the Weight Vector $\boldsymbol{\theta}$} 
\begin{figure}[htbp]
  \centering
  \includegraphics[width=7.5cm]{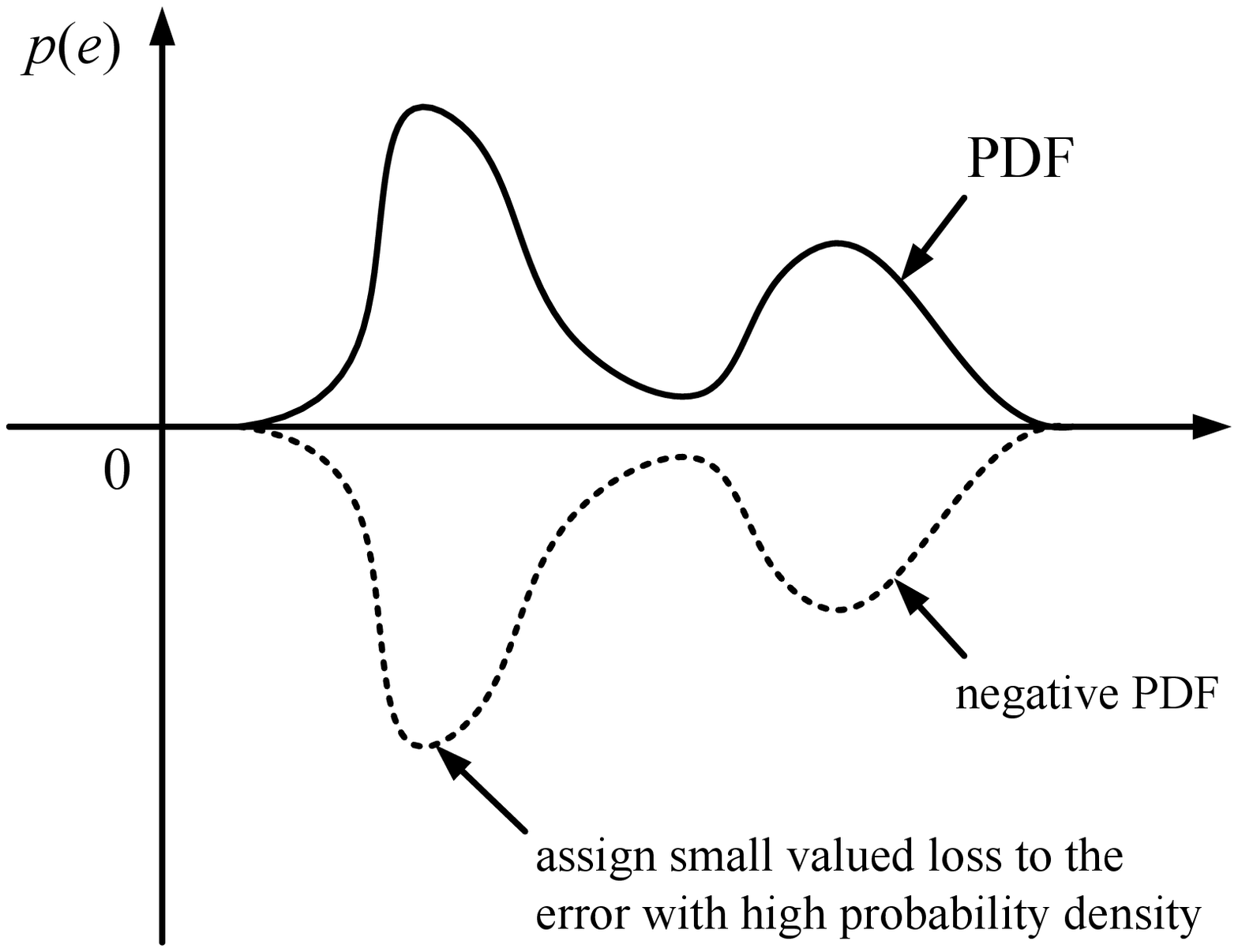}
  \caption{ Error's PDF and negative PDF.}
  \label{fig5}
\end{figure}

Suppose we have obtained $N$ error samples $\left\{ {{e_1}, \cdots, {e_N}} \right\}$ from the first stage learning. To simplify the discussion, we assume that $M$ hidden nodes (radial basis functions) have already been determined (we will discuss this issue later), and we only need to learn the output weight vector $\boldsymbol{\theta}  = {\left[ {{\theta}_1, \cdots ,{\theta _M}} \right]^T}$. In the following, an example of using the popular PDF matching based method is provided. Specifically, we propose to learn the weight vector $\boldsymbol{\theta}$ such that the input-output mapper of the ELN (i.e., the error loss function $l(e)$) is as close as possible to the error's negative PDF $- p(e)$. To further explain why we solve $\boldsymbol{\theta}$ in this way, here we give some explanation. Fig.~\ref{fig5} illustrates an error's PDF (solid) and negative PDF (dotted). If the error loss function is close to the negative PDF $-p(e)$, it will assign small valued loss to the error with high probability density and large valued loss to the error with low probability density. This is in fact very reasonable because the errors at high probability density regions are usually normal errors (we always assume that the normal error is the majority), while the errors at low probability density regions are usually abnormal errors (e.g., errors caused by outliers).
Based on these discussions, we can minimize the following objective function to solve the weight vector $\boldsymbol{\theta}$:
\begin{equation}\label{E8}
\resizebox{0.9\hsize}{!}{$
\begin{array}{l}
J = \int\limits_{ - \infty }^{ + \infty } {{{\left( {l(e) + p(e)} \right)}^2}de} \\
 = \int\limits_{ - \infty }^{ + \infty } {{l^2}(e)de} {\rm{ + 2}}E\left[ {l(e)} \right]{\rm{ + }}\int\limits_{ - \infty }^{ + \infty } {{p^2}(e)de} \\
= \int\limits_{ - \infty }^{ + \infty } {{{\left( {\sum\limits_{j = 1}^M {{\theta _j}{\phi _j}\left( e \right)} } \right)}^2}de} {\rm{ + 2}}E\left[ {\sum\limits_{j = 1}^M {{\theta _j}{\phi _j}\left( e \right)} } \right]{\rm{ + }}\int\limits_{ - \infty }^{ + \infty } {{p^2}(e)de},
\end{array}
$}
\end{equation}
where $E\left[ . \right]$ is the expectation operator. Since the term $\int\limits_{ - \infty }^{ + \infty } {{p^2}(e)de} $ is independent of $\boldsymbol{\theta}$, we have
\begin{equation}\label{E9}
\resizebox{0.9\hsize}{!}{$
\begin{array}{l}
{\boldsymbol{\theta}^*} = \mathop {\arg \min }\limits_{\boldsymbol{\theta}  \in {\mathbb{R}^M}} \left\{ {\int\limits_{ - \infty }^{ + \infty } {{{\left( {\sum\limits_{j = 1}^M {{\theta _j}{\phi _j}\left( e \right)} } \right)}^2}de} {\rm{ + 2}}E\left[ {\sum\limits_{j = 1}^M {{\theta _j}{\phi _j}\left( e \right)} } \right]} \right\}\\
{\rm{   }} \ \ \ = \mathop {\arg \min }\limits_{\boldsymbol{\theta}  \in {\mathbb{R}^M}} \left\{ {{\boldsymbol{\theta} ^T}\textbf{K}\boldsymbol{\theta} {\rm{ + 2}}{\boldsymbol{\theta} ^T}\boldsymbol{\xi} } \right\},
\end{array}
$}
\end{equation}
where $\mathbf{K}$ is an $M \times M$ matrix with ${\textbf{K}_{ij}} = \int_{ - \infty }^{ + \infty } {{\phi _i}\left( e \right){\phi _j}\left( e \right)de}$, and $\boldsymbol{\xi}  = {\left[ {E\left[ {{\phi _{\rm{1}}}\left( e \right)} \right], \cdots, E\left[ {{\phi _M}\left( e \right)} \right]} \right]^T}$. In practice, the vector $\boldsymbol{\xi}$ can be estimated by $\hat {\boldsymbol{\xi}}  = {\left[ {\frac{1}{N}\sum_{i = 1}^N {{\phi _1}\left( {{e_i}} \right)}, \cdots, \frac{1}{N}\sum_{i = 1}^N {{\phi _M}\left( {{e_i}} \right)} } \right]^T}$. Particularly, if the $M$ radial basis functions are assumed to be Gaussian kernel functions with their centers being $c_1, c_2, ...,c_M$ and kernel widths being $\sigma_1, \sigma_2, ..., \sigma_M$, respectively, the vector $\boldsymbol{\xi}$ can be further calculated by $\hat {\boldsymbol{\xi}}  = {\left[ {\frac{1}{N}\sum_{i = 1}^N {G_{\sigma_1}(e_i - c_1)}, \cdots ,\frac{1}{N}\sum_{i = 1}^N {G_{\sigma_M}(e_i - c_M)} } \right]^T}$ and ${\textbf{K}_{ij}}$ has the form of ${\textbf{K}_{ij}}= G_{ \sqrt{{\sigma_i}^2+ {\sigma_j}^2} }(c_i - c_j)$. The calculation of  ${\textbf{K}_{ij}}$ can refer to Appendix A. Thus, one can obtain the following solution:
\begin{equation}\label{E10}
\boldsymbol{\theta} ^{*} =  - {(\textbf{K} + \gamma_{_1} \textbf{I})^{ - {\rm{1}}}}\hat {\boldsymbol{\xi}} ,
\end{equation}
where $\gamma_{_1} \ge 0$ is a regularization parameter to avoid the numerical problem in matrix inversion.

\textit{Remark 1}: The idea of using PDF to design a loss function has already appeared in some existing research works, such as the design of MEE criterion and the design of MMKCC. All of these research works showed that using PDF to design a loss function is an excellent candidate. However, such method, in general, involves more calculation cost, since it needs to train the loss itself. In addition to the PDF matching method discussed in detail in this section, one can also train an ELN with other productive methods, such as the quantized method or some fast heuristic search algorithms \cite{QMEE, PSO2008, WOA}. 

\subsubsection{Determine the $M$ Radial Basis Functions}
There are many types of radial basis functions that can be selected to construct the ELN model, such as the Gaussian kernel, the Laplacian kernel, and some extended Gaussian kernels \cite{GMCC, KRSL, KMPE, CCQKLMS}. In the following, we will focus on the most commonly used Gaussian kernel; however one should note that ${{\phi_i}, i\!=\! 1, 2,\!\cdots\!,M}$ in the ELN model can be selected differently for different nodes to get a better performance.

Following the idea of \cite{RBF1988} and \cite{RBF1989}, when the Gaussian kernel is adopted to construct an RBF network, the centers of $M$ radial basis functions can be directly selected from training samples. Since the proposed ELN shares a similar structure to the traditional RBF neural network, such idea is, of course, suitable to ELN. Therefore, ${\phi _i}(e)$ can be expresssed by
\begin{equation} \label{GaussFun1}
	{\phi _i}(e) \!=\! \frac{1}{\sqrt{2 \pi}\sigma} \exp \left( {  \frac{{{-(e-c_i)^2}}}{{2{{\sigma_i} ^2}}}} \right), i\!=\! 1, 2,\!\cdots\!,M ,
\end{equation}	
where $c_i$ is the center of the $i$-th Gaussian kernel, and $\sigma_i$ is the related kernel width.
In practice, the selection of $\left\{ {{c_1}, \cdots ,{c_M}} \right\}$ does not have to be limited to a fixed manner. For instance, when the number of error samples obtained in the first stage learning is very limited, all the error samples can be selected to construct the ELN model, and when the number of available error samples is very large, the random sampling technology \cite{RS}, the k-means clustering technology \cite{K-means}, and the probability density rank-based quantization technology \cite{PRQ} can be good candidates. In addition, following the idea of randomized learning machines \cite{RVFLN1992, ELM-K, BLS}, it would be promising to generate $\left\{ {{c_1}, \cdots, {c_M}} \right\}$ in a completely random way that is independent of the error samples.

Once $\left\{ {{c_1}, \cdots, {c_M}} \right\}$ are determined, the related $M$ kernel widths $\left\{ {{\sigma_1}, \cdots, {\sigma_M}} \right\}$ should also be properly selected. A simple but useful way to select these kernel widths is to set all kernel widths to the same value, i. e.,  $ {{\sigma_1} = {\sigma_2}=\cdots ={\sigma_M} = {\sigma}} $, and then searching $\sigma$ in a predefined candidate set via cross-validation or other methods. However, since it is difficult to list the potential values of the kernel width exhaustively, the final selected $\sigma$ may not be the desired one. To alleviate
this issue, we propose to set kernel widths in the manner of
\begin{equation} \label{KernelWidth1}
	\sigma_i = {\rm max}(\sigma + n_i,  \epsilon), i\!=\! 1, 2,\!\cdots\!,M ,
\end{equation}	
where $\sigma$ is a reference value for all kernel widths, $n_i$ is a small perturbation term which is assumed to be drawn from Gaussian distribution with zero-mean and variance $\epsilon$, and ``${\rm max}$" denotes the maximum value operator.

\subsection{Training the Learning Machine}
Now we get into the second stage of the learning, i.e., training the learning machine using the obtained error loss function $l(e) = \sum\limits_{j = 1}^M {\theta _j^*{\phi _j}\left( e \right)}$. For a general case, the input-output mapper of the learning machine can be denoted by ${y_i} = f\left( {{\mathbf{x}_i},\boldsymbol{\beta}} \right)$, where $\boldsymbol{\beta}  = {\left[ {{\beta_1}, \cdots ,{\beta_K}} \right]^T} \in {\mathbb{R}^K}$ is a $K$-dimensional weight vector to be learned. In this case, the training of the learning machine can be formulated as the following optimization:
\begin{align}\label{E11}
{\boldsymbol{\beta}^{\rm{*}}} & = \mathop {\arg \min }\limits_{\boldsymbol{\beta}  \in {\mathbb{R}^K}} L(\boldsymbol{\beta})\nonumber\\
&=\mathop {\arg \min }\limits_{\boldsymbol{\beta}  \in {\mathbb{R}^K}} \left\{ {\frac{1}{N}\sum\limits_{i = 1}^N {\sum\limits_{j = 1}^M {\theta _j^*{\phi _j}\left( {{e_i}} \right)} }  + \frac{\gamma_{_2}}{2} {{\left\| \boldsymbol{\beta}\right\|}^2}} \right\},
\end{align}
where ${e_i} = d_{i} - f\left( {{\mathbf{x}_i},\boldsymbol{\beta} } \right)$, $L(\boldsymbol{\beta}) = \frac{1}{N}\sum\nolimits_{i = 1}^N {\sum\nolimits_{j = 1}^M {\theta _j^*{\phi _j}\left( {{e_i}} \right)} }  + \frac{\gamma_{_2}}{2} {\left\| \boldsymbol{\beta}  \right\|^2}$, and $\gamma_{_2} \ge 0$ is a regularization parameter. The above optimization problem is in general nonconvex and there is no closed-form solution. One can, however, use a gradient based method to search the solution. Below we show that if the mapper ${y_i} = f\left( {{\mathbf{x}_i},\boldsymbol{\beta} } \right)$ is an LIP model (such as the Gaussian kernel model \cite{RBF1988}, functional link neural network \cite{RVFLN1992}, extreme leaning machine \cite{ELM-K}, broad learning system \cite{BLS}, etc.), an efficient fixed-point iterative algorithm is available to solve the solution.

Consider an LIP model whose output is computed by
\begin{align}\label{E12}
{y_i} &= {\mathbf{h}_i}\boldsymbol{\beta} \nonumber\\
&= \left[ {{\varphi _1}({\mathbf{x}_i}),{\varphi _2}({\mathbf{x}_i}), \cdots ,{\varphi _K}({\mathbf{x}_i})} \right]{\left[ {{\beta_1},{\beta_2}, \cdots ,{\beta_K}} \right]^T},
\end{align}
where ${\mathbf{h}_i} = \left[ {{\varphi _1}({\mathbf{x}_i}),{\varphi _2}({\mathbf{x}_i}), \cdots ,{\varphi _K}({\mathbf{x}_i})} \right] \in {\mathbb{R}^K}$ is the nonlinearly mapped input vector (a row vector), with ${\varphi _k}(.)$ being the $k$-th nonlinear mapping function ($k = 1,2, \cdots, K$). Let ${{\partial L(\boldsymbol{\beta} )} \mathord{\left/
 {\vphantom {{\partial L(\omega )} {\partial \omega }}} \right.
 \kern-\nulldelimiterspace} {\partial \boldsymbol{\beta} }} = 0$ and ${\phi _j}\left( e_{i} \right) = G_{\sigma_{j} }\left( e_{i} - c_{j} \right)$, we have
\begin{equation}\label{E13}
\resizebox{0.89\hsize}{!}{$
\begin{array}{l}
\frac{1}{N}\sum\limits_{i \!=\! 1}^N \sum\limits_{j \!= \!1}^M \frac{\theta _{j}^{*}}{\sigma_{j}^2}G_{\sigma_{j}}\left( e_{i} \!-\! c_{j} \right)\left( e_{i} \!-\! c_{j} \right)\mathbf{h}_{i}^{T}  \!+\! \gamma_{_2} \boldsymbol{\beta}\!=\!0 \\
\!\Leftrightarrow\!\sum\limits_{i \!= \!1}^N \sum\limits_{j \!= \!1}^M \frac{\theta _{j}^{*}}{\sigma_{j}^2}G_{\sigma_{j}}\left( e_{i} \!-\! c_{j} \right)\left( d_{i} \!-\! {\mathbf{h}_i}\boldsymbol{\beta} \!- \!c_{j} \right)\mathbf{h}_{i}^{T}  \!+\! \gamma_{_2}^{'} \boldsymbol{\beta}  \!=\! 0 \\
\!\Leftrightarrow\!\sum\limits_{i \!=\! 1}^N \psi(e_{i}) d_{i}\mathbf{h}_{i}^{T} \!- \!\sum\limits_{i \!=\! 1}^N \xi(e_{i}) \mathbf{h}_{i}^{T} \!=\! \sum\limits_{i \!=\! 1}^N \psi(e_{i})\mathbf{h}_{i}^{T}\mathbf{h}_{i}\boldsymbol{\beta}  \!-\! \gamma_{_2}^{'} \boldsymbol{\beta},
\end{array}
$}
\end{equation}
where $\gamma_{_2}^{'} = N\gamma_{_2}$, $\psi(e_{i})=\sum\limits_{j \!= \!1}^M\frac{\theta _{j}^{*}}{\sigma_{j}^2} G_{\sigma_{j}}\left( e_{i} - c_{j} \right)$, and $\xi(e_{i})=\sum\limits_{j \!= \!1}^M\frac{c_{j}\theta _{j}^{*}}{\sigma_{j}^2} G_{\sigma_{j}}\left( e_{i} - c_{j} \right)$.
From \eqref{E13}, one can obtain
\begin{equation}\label{E1401}
	\boldsymbol{\beta} =\textbf{R}(\boldsymbol{\beta})^{-1}\textbf{p}(\boldsymbol{\beta}),
\end{equation}
with
\begin{align}\label{E1402}
	\textbf{R}(\boldsymbol{\beta}) &= \sum\limits_{i = 1}^N \psi(e_{i})\mathbf{h}_{i}^{T}\mathbf{h}_{i} - \gamma_2' \mathbf{I}\nonumber\\
	& = \mathbf{H}^{T} \boldsymbol{\Lambda} \mathbf{H} - \gamma_2' \mathbf{I},
\end{align}
and
\begin{align}\label{E1403}
	\textbf{p}(\boldsymbol{\beta}) &= \sum\limits_{i = 1}^N \psi(e_{i}) d_{i}\mathbf{h}_{i}^{T} - \sum\limits_{i = 1}^N \xi(e_{i}) \mathbf{h}_{i}^{T}\nonumber\\
	& = \mathbf{H}^{T} \boldsymbol{\Lambda} \mathbf{d} - \mathbf{H}^{T}\boldsymbol{\vartheta}.
\end{align}
where $\textbf{H}=[{\mathbf{h}_1}^T, {\mathbf{h}_2}^T, \cdots,{\mathbf{h}_N}^T]^T$, $\textbf{d}=[d_{1}, d_{2}, \cdots, d_{N}]^T$, $\boldsymbol{\vartheta}=[\xi(e_{1}), \xi(e_{2}), \cdots, \xi(e_{N})]^T$, and $\boldsymbol{\Lambda}$ is a diagonal
matrix with entries $\boldsymbol{\Lambda}_{i,i}=\psi(e_{i}), i=1,2,\cdots, N$.

It should be noted that both $\textbf{R}(\boldsymbol{\beta})$ and $\textbf{p}(\boldsymbol{\beta})$ in \eqref{E1401} are the functions of $\boldsymbol{\beta}$, and hence it is actually a fixed-point equation that can be described by
\begin{equation}\label{E15}
  \boldsymbol{\beta}  = f_{FP}(\boldsymbol{\beta} ),
\end{equation}
with
\begin{align}\label{E16}
  f_{FP}(\boldsymbol{\beta} )  = \textbf{R}(\boldsymbol{\beta})^{-1}\textbf{p}(\boldsymbol{\beta}).
\end{align}
Such fixed-point equation can be solved by the popular fixed-point iterative method \cite{FP-Theory, FP-MCCT, FP-MCC-MEE} and the details are shown in Algorithm 1.

\begin{algorithm}[htp]\label{Alg1}
\caption{Fixed-point Iterative Algorithm with ELN}
\begin{algorithmic}
\STATE{\rule{-0.5cm}{0.4cm}}\textbf{Input}: training set $\{\textbf{x}_i, y_i\}_{i=1}^{N}$.
\STATE{\rule{-0.5cm}{0.4cm}}\textbf{Output}: weight vector $\boldsymbol{\beta}$.
\STATE{\rule{-0.3cm}{0.4cm}}1. \textbf{Parameters setting}: number of radial basis functions : \\ $M$; reference value for all kernel widths: $\sigma$; variance of perturbation term: $\epsilon$;
regularization parameters: $\gamma_1$, $\gamma_2'$; maximum number of iterations: $T$; tolerance: $\tau$. 
\STATE{\rule{-0.3cm}{0.4cm}}2. \textbf{Initialization}: set  $\boldsymbol{\beta}(0)\!=\!\textbf{0}$ and the $M$ kernel widths following \eqref{KernelWidth1}; meanwhile, construct ${\mathbf{h}_i}$ following a 
\STATE{\rule{-0.3cm}{0.4cm}}  \ \ preset LIP model.\\
\STATE{\rule{-0.3cm}{0.4cm}}3. \textbf{for} $t = 1, ...,T$ \textbf{do}
\STATE{\rule{-0.3cm}{0.4cm}}4. \ \ \  Compute the outputs: ${y_i}\! =\! {\mathbf{h}_i}\boldsymbol{\beta}(t\!-\!1), i\!=\!1,2,\!\cdots\!,N$.
\STATE{\rule{-0.3cm}{0.4cm}}5. \ \ \ Compute the errors: ${e_i}\! =\! {d_i} - {y_i}, i\!=\!1,2,\!\cdots\!,N$.
\STATE{\rule{-0.3cm}{0.4cm}}6. \ \ \ Determine the $M$ centers $\left\{ {{c_1}, \cdots, {c_M}} \right\}$ using the 
\STATE{\rule{-0.3cm}{0.4cm}} \ \ \ \ \ random sampling technology or other methods.
\STATE{\rule{-0.3cm}{0.4cm}}7. \ \ \ Compute $\textbf{K}$: ${\textbf{K}_{ij}}= G_{ \sqrt{{\sigma_i}^2+ {\sigma_j}^2} }(c_i - c_j)$.
\STATE{\rule{-0.3cm}{0.4cm}}8. \ \ \ Compute $\hat {\boldsymbol{\xi}}$: $\hat {\boldsymbol{\xi}}  = [\hat {\boldsymbol{\xi}}_1, \cdots, \hat {\boldsymbol{\xi}}_m,\cdots,\hat {\boldsymbol{\xi}}_M]$, where 
\STATE{\rule{-0.3cm}{0.4cm}} \ \ \ \ \ ${\boldsymbol{\xi}}_m = \frac{1}{N}\sum_{i \!=\! 1}^N {G_{\sigma_m}(e_i \!-\! c_m)}$.
\STATE{\rule{-0.3cm}{0.4cm}}9. \ \ \ Compute  $\boldsymbol{\theta}^{*}$: $\boldsymbol{\theta} ^{*} =  - {(\textbf{K} + \gamma_{_1} \textbf{I})^{ - {\rm{1}}}}\hat {\boldsymbol{\xi}}$.
\STATE{\rule{-0.3cm}{0.4cm}}10. \ \ Compute $\boldsymbol{\Lambda}$: $\boldsymbol{\Lambda}\!=\! \text{diag}[\psi(e_{1}),\!\cdots\!, \psi(e_{i}),\!\cdots\!,\psi(e_{N})]$, \STATE{\rule{-0.3cm}{0.4cm}} \ \ \ \ \ \  where $\psi(e_{i})=\sum\nolimits_{j \!= \!1}^M\frac{\theta _{j}^{*}}{\sigma_{j}^2} G_{\sigma_{j}}\left( e_{i} - c_{j} \right)$.
\STATE{\rule{-0.3cm}{0.4cm}}11. \ \ Compute $\boldsymbol{\vartheta}$:  $\boldsymbol{\vartheta}=[\xi(e_{1}), \cdots,\xi(e_{i}), \cdots, \xi(e_{N})]^T$, \STATE{\rule{-0.3cm}{0.4cm}} \ \ \ \ \ \  where $\xi(e_{i})=\sum\nolimits_{j \!= \!1}^M\frac{c_{j}\theta _{j}^{*}}{\sigma_{j}^2} G_{\sigma_{j}}\left( e_{i} - c_{j} \right)$.
\STATE{\rule{-0.3cm}{0.4cm}}12. \ \ Compute $\textbf{H}$: $\textbf{H}=[{\mathbf{h}_1}^T, {\mathbf{h}_2}^T, \cdots,{\mathbf{h}_N}^T]^T$.
\STATE{\rule{-0.3cm}{0.4cm}}13. \ \  Update $\boldsymbol{\beta}$: $\boldsymbol{\beta}(t) \!=\!\big(\mathbf{H}^{T} \boldsymbol{\Lambda} \mathbf{H} \!-\! \gamma_2' \mathbf{I}\big)^{\!-\!1}  \big(\mathbf{H}^{T} \boldsymbol{\Lambda} \mathbf{d} \!-\! \mathbf{H}^{T}\boldsymbol{\vartheta}\big)$.
\STATE{\rule{-0.3cm}{0.4cm}}14. \ \  Until $\|\boldsymbol{\beta}(t)\!-\!\boldsymbol{\beta}(t-1)\|^{2}/\|\boldsymbol{\beta}(t-1)\|^{2}\!<\!\tau$.
\STATE{\rule{-0.3cm}{0.4cm}}15. \textbf{end for}
\end{algorithmic}
\end{algorithm}

\textit{Remark 2}: Algorithm $1$ is mainly designed for the case of multiple-input-single-output. To extend it to multiple-input-multiple-output cases, one can directly refer to the method proposed in \cite{RFFEXKRLS}.

\section{Experimental Results} \label{Sec4}
In this section, we present experimental results to demonstrate the performance of the proposed method. Except mentioned otherwise,
all results are obtained using MATLAB (R2016b) on a
machine equipped with Intel Xeon E3-1505M V6 CPU and 16-GB
RAM.

\subsection{Performance Evaluation on Regression Tasks}\label{SecRegression}

\subsubsection{Linear Regression}\label{SecLinearRegression}
To test the performance of the proposed method in the scenario of linear regression, the weight vector of a linear system is assumed to be $\boldsymbol{\beta}^* = [2, 1]^T$, and the input vectors $\textbf{x}_i, i=1,2,\cdots,N$ of the system are assumed to be uniformly distributed over $[-2,2]\times[-2,2]$. Hence, the system outputs can be expressed by
\begin{equation}\label{ESysetmIdentification0}
	y_i = \textbf{x}_i^T{\boldsymbol{\beta}^*} , i=1,2,\cdots,N .
\end{equation}
An additional interference sequence is then added to the system outputs to model the effect of noise and outliers, i. e.,
\begin{equation}\label{ESysetmIdentification1}
	d_i = y_i + v_i, i=1,2,\cdots,N,
\end{equation}
where $d_i$ is the noisy observation of the $i$-th output of the system. Our goal is to find the weight vector of the system based on $N$ pairs of observed samples $\{(\textbf{x}_i, d_N),\cdots,(\textbf{x}_i, d_N)\}$. In the experiments, $N$ is set to $500$, and the interference sequence $v_i, i=1,2,\cdots,N$ is modelled as
\begin{equation}\label{ESysetmIdentification2}
	v_i = (1-\eta_i) A_i + \eta_i B_i, i=1,2,\cdots,N,
\end{equation}
where $\eta_i$ is a binary variable with probability mass $Pr(\eta_i=1) = p $ and $Pr(\eta_i=0) = 1- p $; $A_i$ and $B_i$ are two processes to model the inner noise and outliers, respectively.
Specifically, we set $p = 0.1$, and set $B_i$ to be a Gaussian process with zero-mean and variance $100$. For the distribution of $A_i$, four cases are considered: 1) symmetric Gaussian mixture density: $1/2\mathcal{N}(-5, 0.1) + 1/2\mathcal{N}(5, 0.1)$, where $\mathcal{N}(\mu, \delta)$ denotes a Gaussian density function with mean $\mu$ and variance $\delta$;  2) asymmetric Gaussian mixture density: $1/3\mathcal{N}(-3, 0.1) + 2/3\mathcal{N}(5, 0.1)$;  3) single Gaussian density: $\mathcal{N}(0, 0.1)$; 4) uniform distribution over $[0, 1]$.
To measure the performance of the proposed method, the root mean square deviation (RMSD) is defined by
\begin{equation}\label{ESysetmIdentification4}
	{\rm RMSD} = \sqrt{\frac{1}{2}\|\hat{\boldsymbol{\beta}} -\boldsymbol{\beta}^*\|^2},
\end{equation}
where $\hat{\boldsymbol{\beta}}$ is the estimate of $\boldsymbol{\beta}^*$.

\begin{table*}[htbp]
	\renewcommand\arraystretch{1.3}
	\newcommand{\tabincell}[2]{\begin{tabular}{@{}#1@{}}#2\end{tabular}}
	\caption{Performance Comparison of Different Methods on Linear Regression Averaged Over $100$ Independent Runs (The best results are marked in bold, the second best results are underlined, and the symbol N/A means that the corresponding value is less than 0.0001)}
	\begin{center}
		\begin{tabular}{llccccccccccc}
			\toprule
			Index &Noise Type  & MSE & MCC & MMCC & MCC-VC & GMCC & KRSL & KMPE & QMEE & ELN1 & ELN2\\
			\midrule
			\multirow{1}{*}{RMSD}
				&Case 1&0.1686&0.1662&0.1659&0.0319&0.0922&0.0899&0.1010&0.0148&\textbf{0.0116}&\underline{0.0139}\\
				&Case 2&0.1667&0.1619&0.1612&0.0247&0.0817&0.0850&0.0903&\underline{0.0145}&\textbf{0.0125}&\underline{0.0145}\\
				&Case 3&0.1133&0.0131&0.0131&0.0131&0.0131&\underline{0.0130}&0.0138&0.0133&0.0164&\textbf{0.0125}\\
				&Case 4&0.1143&0.0193&0.0187&\textbf{0.0112}&0.0163&0.0147&0.0163&0.0115&\underline{0.0119}&0.0121\\ 		
			\midrule
			\multirow{1}{*}{Time}
				&Case 1&\textbf{N/A}&\underline{0.0006}&0.0008&0.0063&0.0169&0.0134&0.0103&0.0318&0.0570&0.2094\\
				&Case 2&\textbf{N/A}&\underline{0.0006}&\underline{0.0006}&0.0067&0.0282&0.0216&0.0128&0.0653&0.0407&0.2282\\
				&Case 3&\textbf{N/A}&\underline{0.0008}&0.0012&\underline{0.0008}&0.0025&0.0044&0.0043&0.0346&0.0587&0.1653\\
				&Case 4&\textbf{N/A}&\underline{0.0012}&\underline{0.0012}&0.0026&0.0061&0.0054&0.0047&0.0346&0.0310&0.3158\\
			\bottomrule
		\end{tabular}
		\label{TabSystem}
	\end{center}
\end{table*}

First, the performance comparison among ELN, MSE, MCC, MMCC, MCC-VC, GMCC, KRSL, KMPE, and QMEE is carried out, and the results are shown in Table~\ref{TabSystem}. Herein, both the random sampling technology \cite{RS} and the probability density rank-based quantization technology \cite{PRQ} are considered to generate the centers of the ELN model. For the convenience of distinction, they are marked as ELN1 and ELN2, respectively. To guarantee a fair comparison, the parameters for each method are chosen by performing a grid search and measuring
performance by an additional stratified ten-fold cross validation on
the training data\footnote{We choose the one group parameters that can achieve minimum average value of the sum of squares of validation errors.}. In detail, the parameter search ranges of different methods are as follows.

\begin{itemize}
	\item  For MSE, the regularization parameter is searched from $\{10^{-5},10^{-4},\cdots,10^{4},10^{5}\}$.
	\item  For MCC, its regularization parameter is searched in the same range of MSE, and its kernel width is searched from $\{0.1,  0.3,  0.5,  0.7,  1, 3, 5, 7, 10, 15, 30, 60, 100\}$.
	\item  For MMCC, the regularization parameter is searched in the range of $\{10^{-5},10^{-4},\cdots,10^{4},10^{5}\}$, and the parameters $\sigma_1$, $\sigma_2$ and $\lambda$ are chosen from  $\{0.1, 0.3, 0.5, 0.7, 1, 3, 5, 7\}$, $\{1, 3, 5, 7, 10, 15, 30, 100\}$, and $\{0, 0.1, 0.3, 0.5, 0.7, 0.9, 1.0\}$, respectively.
	\item  For MCC-VC, its center of Gaussian kernel is searched from $\{-5, -3, -1, -0.5, 0, 0.5, 1, 3, 5\}$, while the search ranges of the remaining parameters are kept the same of MCC.
	\item  For GMCC, the regularization parameter is chosen from $\{10^{-5}, 10^{-4}, \cdots, 10^{4}, 10^{5}\}$, the parameter $\alpha$ is searched from $\{1, 2, 3, 4, 5\}$, and the parameter $\lambda$ is searched from $\{0.0005, 0.001, 0.005, 0.01, 0.05, 0.1, 0.5, 1, 5\}$.
	\item  For KRSL, the regularization parameter is chosen from $\{10^{-5},10^{-4},\cdots,10^{4},10^{5}\}$, the parameter $\lambda$ is searched from $\{0.1, 0.3, 0.5, 0.7, 1, 3, 5, 7, 10\}$, and the kernel width is searched from $\{0.1,  0.3,  0.5,  0.7,  1, 3, 5, 7, 10, 15, 30, 60, 100\}$.
	\item  For KMPE, the regularization parameter and the kernel width are searched in the same ranges of KRSL, and the parameter $p$ is searched from $\{0.1, 0.5, 1.0, 1.5, 2.0, 2.5, 3.0, 4.0, 5.0\}$.
	\item  For QMEE, all the parameter search ranges of MCC are shared with it. In addition, the parameter for performing quantization operation in it is set to $0.5$.
	\item  For ELN1 and ELN2, the regularization parameter $\gamma_2'$ is searched from $\{10^{-5},10^{-4},\cdots,10^{4},10^{5}\}$, the reference value $\sigma$ for all kernel widths is searched from $\{0.1,  0.3,  0.5,  0.7,  1, 3, 5, 7, 10, 15, 30, 60, 100\}$, the number of radial basis functions is fixed at $M =50$, the regularization parameter $\gamma_1$ is set to $10^{-3}$, and the variance of perturbation term is set as $\epsilon=0$.
\end{itemize}
In addition, for all iteration-based methods, the maximum number of iteration is set as $T=50$, and the tolerance for early stopping is set to  $\tau=10^{-7}$. Based on the data shown in Table~\ref{TabSystem}, the following results can be obtained.
\begin{itemize}
	\item Although MCC, MMCC, MCC-VC, GMCC, KRSL, KMPE, QMEE, ELN1, and ELN2 need more training time in comparison with MSE criterion, they can outperform the latter in terms of RMSD, significantly. This demonstrates the effectiveness of these ITL criteria and the ELN model in robust learning.
	\item  The RMSDs of MMCC, MMCC, MCC-VC, GMCC, KRSL, KMPE, QMEE, ELN1, and ELN2 are smaller than that of MCC on the most of cases. This demonstrates that the criterion designed based on a single Gaussian kernel with zero-mean and fixed variance can be improved by changing its mean, variance, kernel type, or by integrating the advantages of different kernels.
	\item  The QMEE, ELN1, and ELN2 can, in general, obtain a better performance in terms of RMSD in comparison with other robust criteria. The reason behind this may be that more kernels are involved in these three criteria, such that they can characterize complex noise distributions better \cite{QMEE}.
	\item  The ELN1 and ELN2 have the ability to obtain a comparative and even smaller RMSD than QMEE. A possible reason for this is that QMEE is built with a simple online-vector quantization (OVQ) technology \cite{QMEE}. Although this technology also chooses the centers of kernels from error samples, it however can not keep the distribution information of the original samples well \cite{PRQ}. Moreover, in the process of performing quantization operator, the contribution of a discarded error sample is directly measured by its closest center, which ignores the difference between two adjacent error samples. In contrast, the proposed ELN model is built based on the idea of RBF networks, whose centers do not depend on any specific manner. Once an appropriate sampling technology, such as the random sampling technology adopted in ELN1 or the probability density rank-based quantization technology in ELN2, is adopted, it can capture more distribution information hidden in the original error samples. Meanwhile, the combination coefficient in the ELN model are determined by the idea of PDF matching, which can, to some extent,  allocate the coefficients of different kernels in a more reasonable manner.
\end{itemize}

Then, the influence of the free parameters on the learning performance of ELN is investigated. In particular, when the radial basis function is assumed to be the Gaussian kernel, the extra parameters in comparison with MMSE criterion to control the learning performance of an ELN include the regularization parameter $\gamma_1$, the number of radial basis functions $M$, the reference value $\sigma$ for all kernel widths, and the variance of perturbation term $\epsilon$. In the following experiments, we investigate the influence of these parameters on the learning performance of ELN in three steps.
\begin{figure*}[htbp]
	\centering
	\subfigure[Case 1]{
		\label{Influence_fig6a} 
		\includegraphics[width=3.3in]{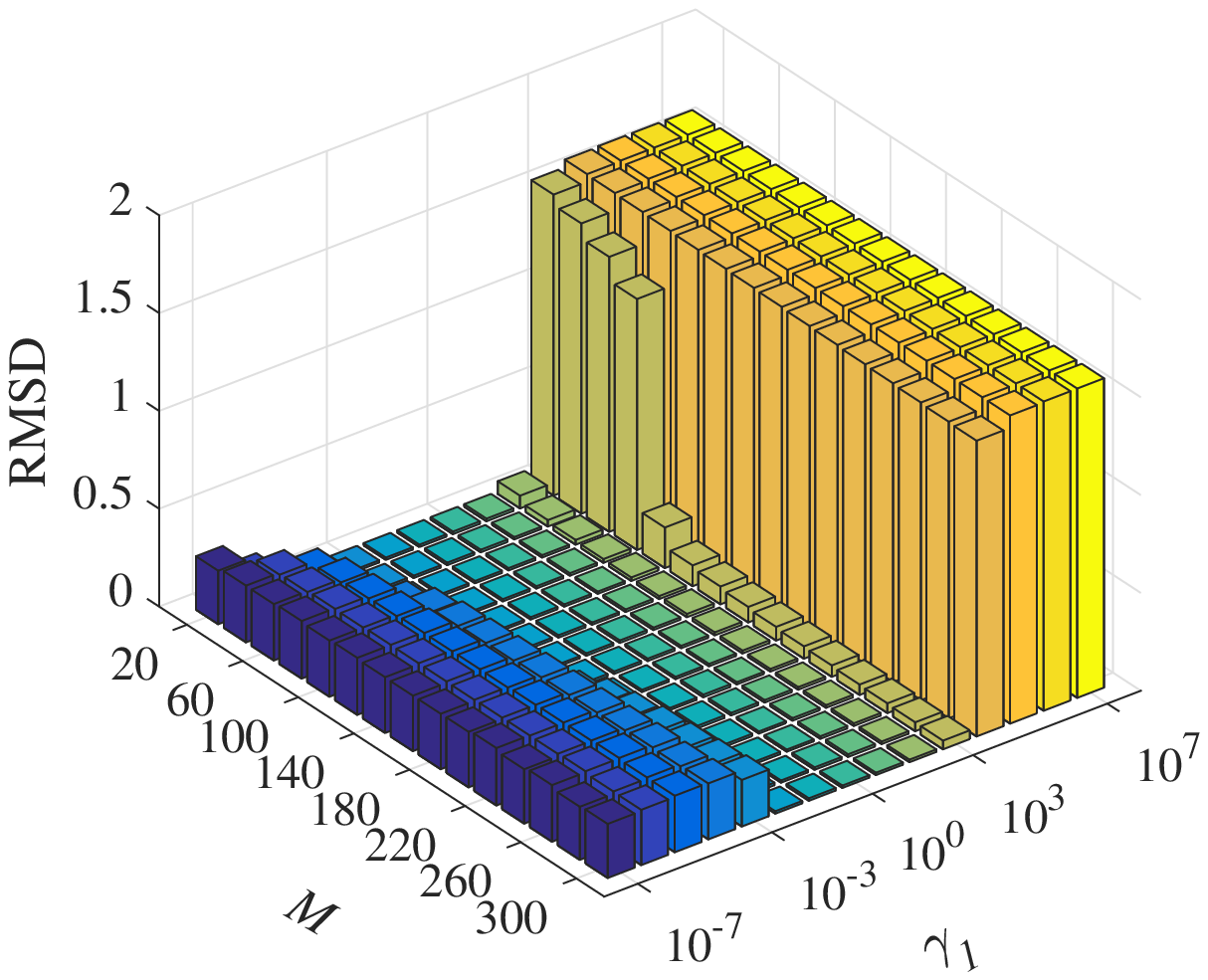}}
	\subfigure[Case 2]{
		\label{Influence_fig6b} 
		\includegraphics[width=3.3in]{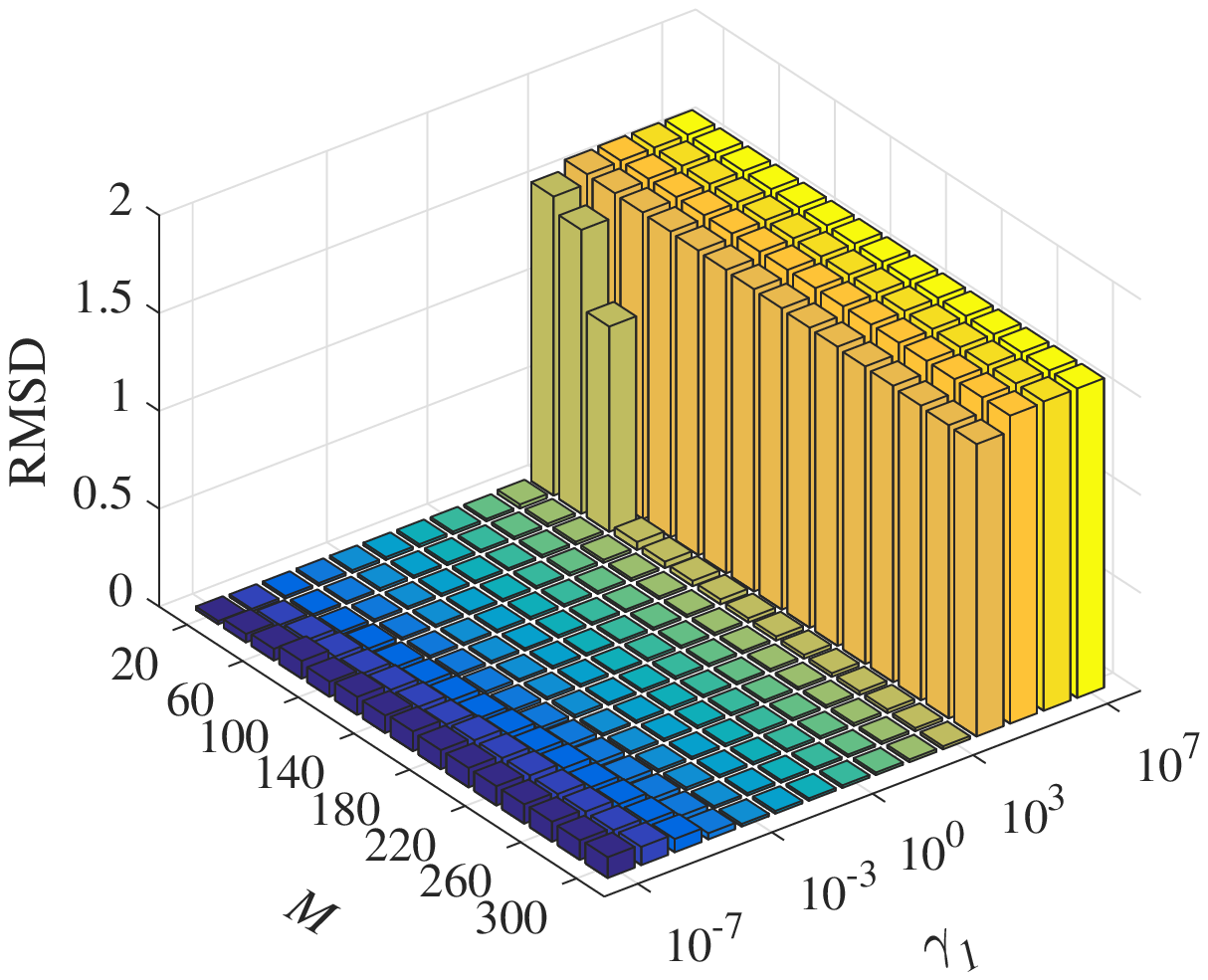}}
	\subfigure[Case 3]{
		\label{Influence_fig6c} 
		\includegraphics[width=3.3in]{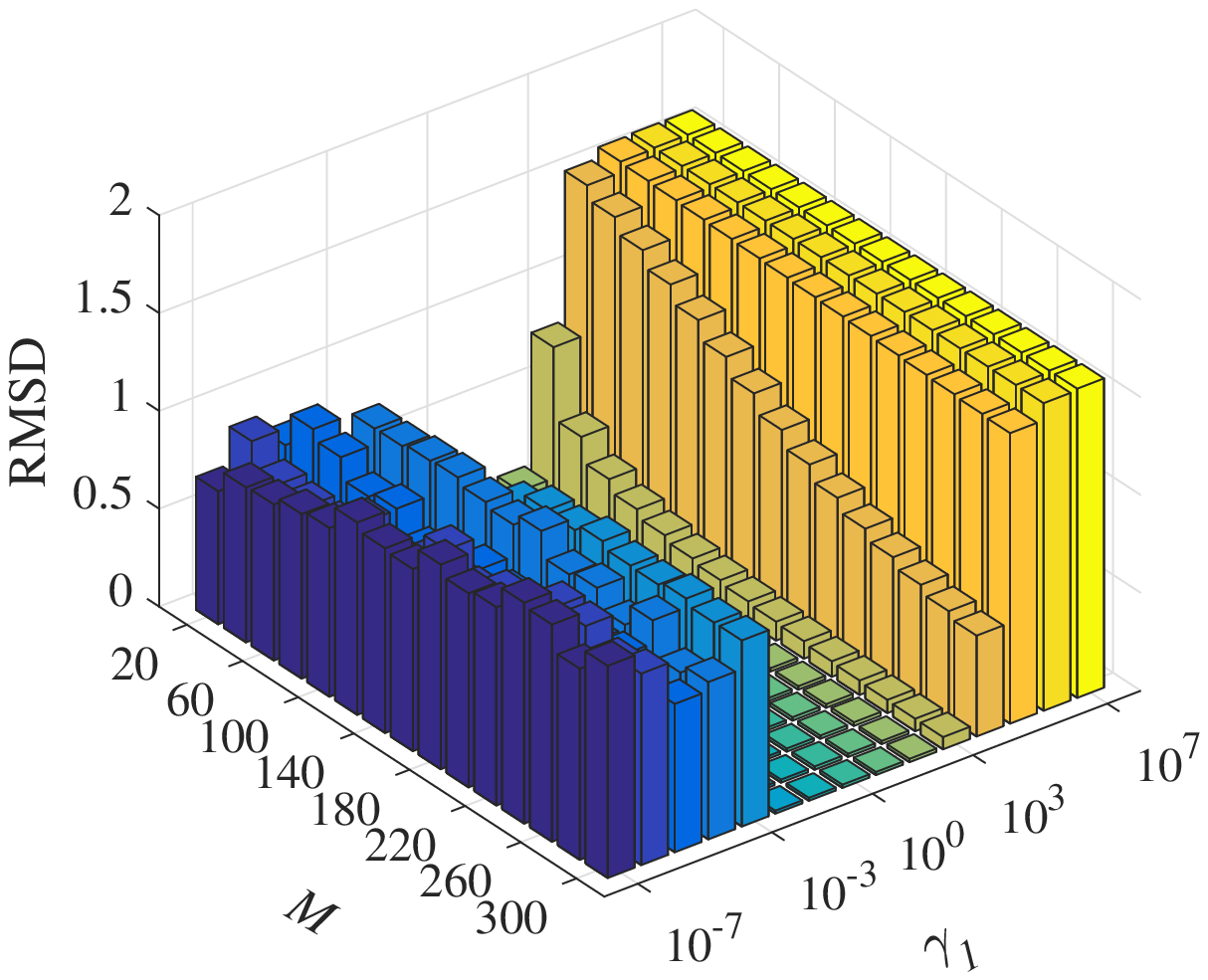}}
	\subfigure[Case 4]{
		\label{Influence_fig6d} 
		\includegraphics[width=3.3in]{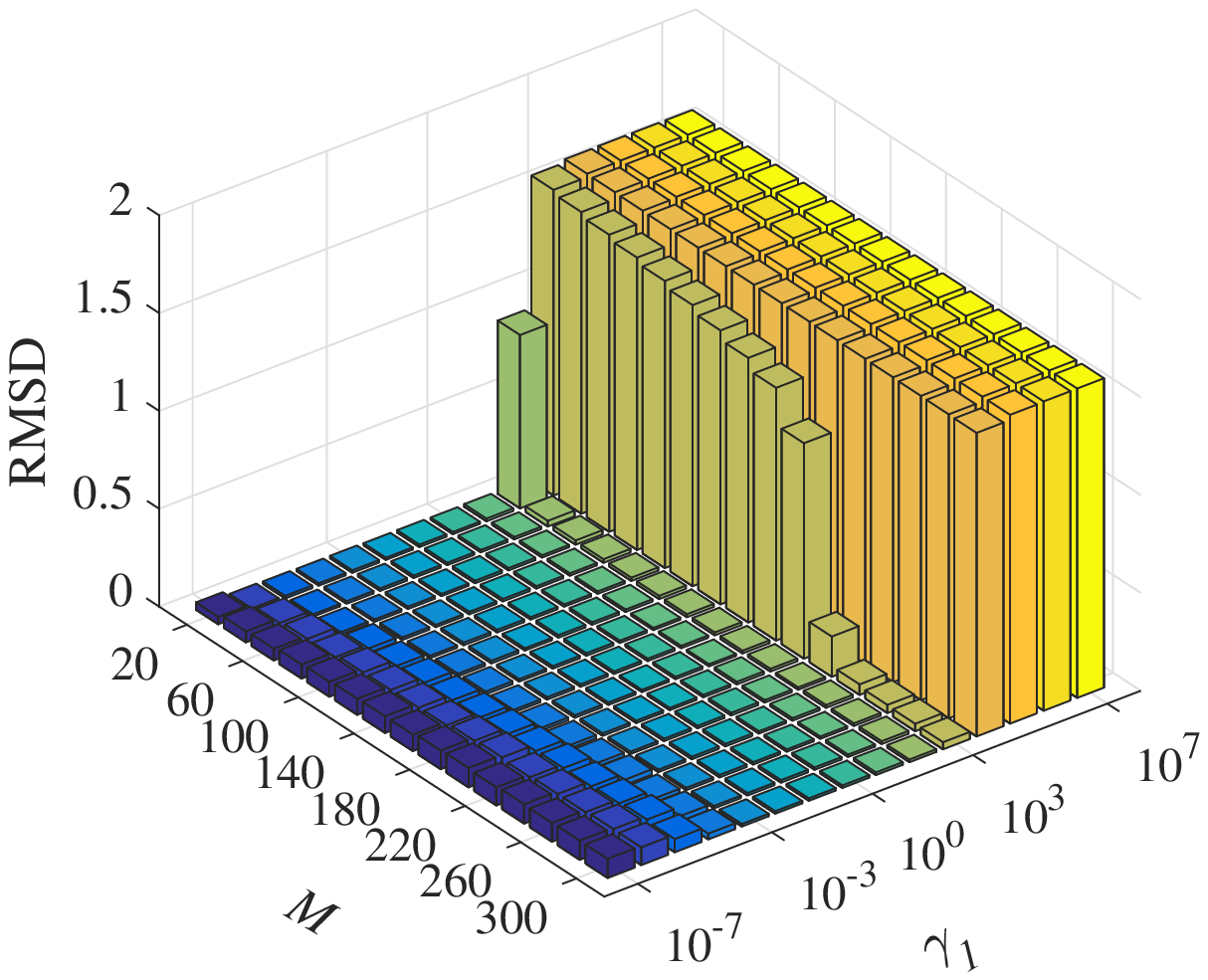}}
	\caption{Influence of the parameters $M$ and $\gamma_1$ on the learning performance of ELN under different noise environments.}
	\label{Influence_fig6}
\end{figure*}
First, on the premise of setting the values of the other parameters to be the same as the previous ones\footnote{For all cases, $\epsilon = 0$ is a default choice. In addition, the values of $\sigma$ for Case 1, Case2, Case 3, and Case 4 are set to $1$, $0.7$, $3$, and $0.7$, respectively; the values of $\gamma_2'$ for Case 1, Case2, Case 3, and Case 4 are set to $10^{-1}$, $10^{-1}$, $10^{-2}$, and $10^{0}$, respectively.}, Fig.~\ref{Influence_fig6} shows the variation of RMSD versus $M$ and $\gamma_1$ under different noise environments.
It can be seen from Fig.~\ref{Influence_fig6} that, although the choices of $M$ and $\gamma_1$ can affect the learning performance of ELN, there is always a large flat area that ELN can obtain a small RMSD close to the optimal value. Moreover, once the value of $\gamma_1$ is appropriately chosen and the value of $M$ exceeds a certain value (for example, $10^{-2}\le\gamma_1 \le10^{2}$ and $M\ge50$), the change of $M$ does not affect the final learning performance of ELN, evidently.
These results suggest that $M$ and $\gamma_1$ are two parameters which can be easy to be tuned, and hence we do not need to pay much attention to the selection of them in practice.
Then, by setting the values of the other parameters to be the same of the previous ones, Fig.~\ref{Influence_fig7} shows the variation of RMSD versus $\sigma$ under different noise environments. As can be seen from Fig.~\ref{Influence_fig7}, the change of $\sigma$ can affect the learning performance of ELN, significantly. In particular, when the value of $\sigma$ is set near $2^0$, the RMSD of ELN can approach its performance shown in Table~\ref{TabSystem}. However, when the value of $\sigma$ is set to be very large or very small, the RMSD of ELN can increase dramatically.
Therefore, $\sigma$ is a parameter that should be carefully chosen in practice.
Other than the grid search method adopted in this paper, other productive technologies, like the whale optimization algorithm (WOA) \cite{WOA} method, can also be a good candidate.
\begin{figure}[htbp]
	\centering
	\includegraphics[width=3.4in]{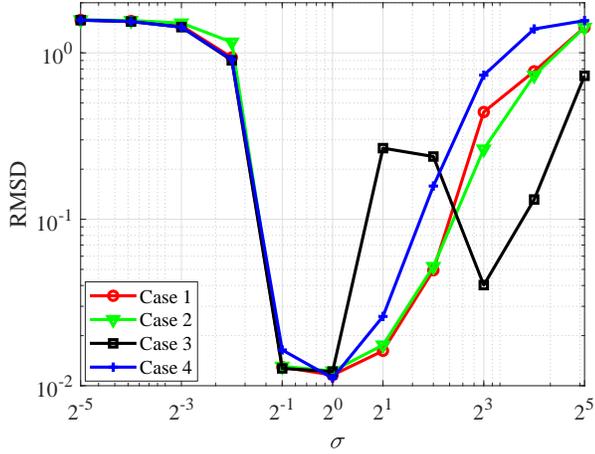}
	\caption{Influence of the parameter $\sigma$ on the learning performance of ELN under different noise environments.}
	\label{Influence_fig7}
\end{figure}
Finally, the influence of $\epsilon$ on the learning performance of ELN is investigated, and the related experimental results are shown in Table~\ref{Influence_EPSILON}.
It can be seen from Table~\ref{Influence_EPSILON} that, instead of always setting $\epsilon = 0$, properly adjusting the value of $\epsilon$ can further enhance the learning performance of ELN. However, it should be noted that the RMSD of ELN will increase dramatically when the value of $\epsilon$ is set to be greater than $10^0$. Thus, in the following experiments, only smaller values of $\epsilon$ will be considered as candidate set.
\begin{table}[htbp]
	\renewcommand\arraystretch{1.1}
	\newcommand{\tabincell}[2]{\begin{tabular}{@{}#1@{}}#2\end{tabular}}
	\caption{Influence of the parameter $\epsilon$ on the learning performance of ELN under different noise environments}
	\begin{center}
		\begin{tabular}{lcccc}
			\toprule
			\multirow{2}{*}{$\epsilon$}  & \multicolumn{4}{c}{RMSD} 	\\
			\cline{2-5}
			& Case 1  & Case 2 & Case 3 & Case 4  \\
			\midrule
			$0$&0.0116&\textbf{0.0125}&0.0164&0.0119\\
			$10^{-7}$&0.0117&0.0132&0.0172&\textbf{0.0110}\\
			$10^{-5}$&\textbf{0.0113}&0.0127&0.0169&0.0121\\
			$10^{-3}$&0.0125&0.0141&0.0152&0.0128\\
			$10^{-1}$&0.3911&2.3767&\textbf{0.0126}&1.7032\\
			$10^{0}$&0.0114&0.0133&0.1234&0.0120\\
			$10^{1}$&0.4792&0.3405&0.0531&0.9983\\
			$10^{3}$&1.5811&1.5811&1.5811&1.5811\\
			$10^{5}$&1.5811&1.5811&1.5811&1.5811\\
			\bottomrule
		\end{tabular}
		\label{Influence_EPSILON}
	\end{center}
\end{table}

\subsubsection{Nonliner Regression}
In this subsection, ten real regression benchmark data sets, taken from UCI machine learning repository \cite{UCI2017}, LIACC\footnote{https://www.dcc.fc.up.pt/\url{~}ltorgo/Regression/DataSets.html} and LIBSVM\footnote{https://www.csie.ntu.edu.tw/\url{~}cjlin/libsvmtools/datasets/},  are further adopted to test the performance of the proposed ELN in nonlinear regression examples. The descriptions of these data sets are shown in Table~\ref{UCIData}. In the experiment, the training and testing
samples for each data set are randomly selected and
the data have been normalized into the range of $[0, 1]$. In addition, an additional noise sequence has been added to the outputs of training samples. The details about the noise model can refer to~\eqref{ESysetmIdentification2}. In particular, we set $p = 0.1$, and set the density functions of $A_i$ and $B_i$ to be $1/2\mathcal{N}(-0.8, 0.01) + 1/2\mathcal{N}(0.8, 0.01)$ and $\mathcal{N}(0, 10)$, respectively.
\begin{table}[htbp]
	\renewcommand\arraystretch{1.05}
	\newcommand{\tabincell}[2]{\begin{tabular}{@{}#1@{}}#2\end{tabular}}
	\caption{Specification of Regression Benchmark Data Sets}
	\begin{center}
		\begin{tabular}{lccccccc}
			\toprule
			Data Set  & Attributes  &  Training  &  Testing\\
			\midrule
			Concrete Slump Test  & 10 & 52 & 52 \\
			Machine CPU  & 7 & 105 & 104  \\
			Residential Building  & 106 & 186 & 186  \\
			Auto MPG  & 8 & 199 & 199  \\
			Real Estate Valuation  & 7 & 207 & 207 \\
			Concrete Compressive Strength  & 9 & 515 & 515 \\
			Space GA & 7 & 1554 & 1553 \\
			Computer Activity & 21 & 4096 & 4096 \\
			Pole & 48 & 7500 & 7500 \\
			California Housing & 9 & 10320 & 10320 \\
			\bottomrule
		\end{tabular}
		\label{UCIData}
	\end{center}
\end{table}

To perform the nonlinear mapping from inputs to outputs of these data sets, the random vector functional link neural network (RVFLNN) \cite{RVFLN1992} is adopted as an LIP model for training. Following the idea of RVFLN, the ${\textbf{h}_i}$ shown in \eqref{E12} can be constructed by
\begin{equation}\label{EFLNN1}
 {\textbf{h}_i} = \left[ {\textbf{x}_i, | {h_1}({\textbf{x}_i}),{h_2}({\textbf{x}_i}), \cdots ,{h_K}({\textbf{x}_i})}  \right],
\end{equation}
with
\begin{equation}\label{EFLNN2}
	{h_k}({\textbf{x}_i}) = g(\textbf{x}_{i}\textbf{w}_{k} + b_{k}), k=1,2,\dots,K,
\end{equation}
where $K$ is the number of hidden nodes of the network, $\textbf{w}_{k}$ is the randomly generated weights which connect the $k$ hidden node to the input layer, $b_{k}$ is the randomly generated bias of the $k$ hidden node, and $g$ is the activation function. In the following experiments, $K$ is set to $200$ for all compared methods and data sets, $\{\textbf{w}_{k}\}_{k=1}^{K}$ are assumed to be drawn from uniform distribution over $[-1, 1]$, $\{b_{k}\}_{k=1}^{K}$ are assumed to be drawn from uniform distribution over $[0, 1]$, and $g$ is set as the ``sigmoid" function whose expression is
\begin{equation}\label{EFLNN3}
	g(x) = \frac{1}{\text{exp}(-x) + 1}.
\end{equation}

\begin{table*}[htbp]
	\renewcommand\arraystretch{1.05}
    \newcommand{\tabincell}[2]{\begin{tabular}{@{}#1@{}}#2\end{tabular}}
	\caption{Test RMSE of Different Methods on Regression Benchmark Data Sets Averaged Over $20$ Independent Runs}
	\begin{center}
		\begin{tabular}{llcccccccccc}
			\toprule
			 Data Set & MSE & MCC & MMCC & MCC-VC & GMCC & KRSL & KMPE & QMEE & ELN \\
			\midrule
		        Concrete Slump Test&0.2997&0.2541&0.2654&0.2471&0.2381&0.6785&0.2946&\textbf{0.2098}&0.2161\\
			    Machine CPU&0.1524&0.1347&0.1359&0.1330&0.1148&0.1360&0.2493&0.0878&\textbf{0.0761}\\
			    Residential Building&0.1611&0.1525&0.1532&0.1525&0.1524&0.1515&0.1511&0.1110&\textbf{0.0917}\\
			    Auto MPG&0.1938&0.1957&0.1984&0.2120&0.1420&0.1232&0.1589&0.1341&\textbf{0.0934}\\
			    Real Estate Valuation&0.1197&0.1175&0.1216&0.1269&0.0995&0.0899&\textbf{0.0884}&0.1022&0.0983\\
			    Concrete Compressive Strength&0.2304&0.2236&0.1686&0.2236&0.1678&0.1786&0.1840&0.1603&\textbf{0.1443}\\
			    Space GA&0.0592&0.0513&0.0558&0.0513&0.0460&0.0456&0.0570&0.0417&\textbf{0.0411}\\
			    Computer Activity&0.1267&0.1188&0.1125&0.1188&0.0758&0.0756&0.0754&\textbf{0.0456}&0.0471\\
			    Pole&0.2610&0.2546&0.2536&0.2546&0.2372&0.2391&0.2359&0.2235&\textbf{0.2220}\\
			    California Housing&0.1497&0.1482&0.1470&0.1482&0.1361&0.1350&0.1349&0.1304&\textbf{0.1294}\\ 		
			\bottomrule
		\end{tabular}
		\label{TabNonlinearRegression}
	\end{center}
\end{table*}
In order to make a fair comparison, the parameters for each method are still chosen by the grid search method. In detail, the parameter search ranges of MSE, MCC, MMCC, MCC-VC, GMCC, KRSL, and KMPE are directly set to the same of those in Section~\ref{SecRegression}1. For QMEE, except that the parameter for performing quantization operation has been changed to $0.2$, other parameters of it are searched in the same ranges described in Section~\ref{SecRegression}1. For ELN, the random sampling technology is chosen to generate the centers of radial basis functions, the search range of $\gamma_{1}$ has been extended to $\{10^{-5}, 10^{-3}, 10^{-1}\}$, the variance of perturbation term is fixed at $\epsilon=10^{-4}$, while the number of radial basis functions is chosen in the manner of
\begin{equation}\label{EFLNN4}
	M =\left\{ \begin{array}{ll}
		N, & \ \   N<50\\
		50, & \ \ 50\le N< 3000\\
		300, & \ \ N\ge 3000 , \\
	\end{array} \right.
\end{equation}
where $N$ is the number of training samples. In addition, since the optimal weight vector  $\boldsymbol{\beta}^{*}$ is not available in this part, RMSD has been replaced with the root mean square Error (RMSE) for performance evaluation, i. e.,
\begin{equation}\label{EFLNN5}
 	{\rm RMSE} = \sqrt{\frac{1}{N_{te}} \sum\limits_{i=1}^{N_{te}}\left(d_i - \hat d_i \right)^2},
\end{equation}
where $d_i$ and $\hat d_i$ are the real output and its estimate of the $i$-th testing sample\footnote{After training ends, a bias with its value being equal to the mean of training errors has been added to the outputs of the model when QMEE and ELN have been adopted for training. Such operator is common to supervised learning methods under the MEE criterion, and the details can refer to \cite{ITL}. Since QMEE and ELN are closely related to MEE, the outputs of the models trained with them are both added with an additional bias.}, respectively, and $N_{te}$ is the number of testing samples.

Table~\ref{TabNonlinearRegression} shows the test RMSE of different methods on the ten regression benchmark data sets. It can be seen from Table~\ref{TabNonlinearRegression} that, although ELN does not always obtain the minimal RMSE on the ten data sets, it can, at least, outperform other compared methods on the most of data sets in terms of RMSE. This suggests that the RVFLNN developed with ELN can be a good candidate to general regression problems.

\subsection{Performance Evaluation on Classification Tasks}
The good performance of the proposed ELN has been demonstrated in some regression tasks.
In this subsection, its performance will be evaluated on some classification benchmark data sets.
In order to test the robustness of different methods, the learning with noisy labels is considered in the following. In particular, the noisy labels are manually generated by a noise transition matrix $Q$, where $Q_{i,j} = Pr(\hat{t} = j\mid t = i)$ gives that noisy $\hat{t}$ is flipped from clean $t$. For a classification problem involving $C$ categories, a popular structure of $Q$ \cite{Ref_new02} can be formulated by
\begin{equation}\label{QQQ}
\resizebox{0.89\hsize}{!}
{$
{Q} \!=\! \left[ {\begin{array}{*{20}{c}}
{1 - \varepsilon }&\varepsilon &0& \cdots &0& \cdots &0\\
0& \ddots & \ddots & \ddots & \vdots & \ddots & \vdots \\
 \vdots & \ddots &{1 - \varepsilon }&\varepsilon &0& \cdots &0\\
0& \cdots &0& \ddots &\varepsilon & \ddots & \vdots \\
 \vdots & \ddots & \vdots & \ddots &{1 - \varepsilon }& \ddots &0\\
0& \cdots &0& \cdots &0& \ddots &\varepsilon \\
\varepsilon &0& \cdots &0& \cdots &0&{1 - \varepsilon }
\end{array}} \right] \in {\mathbb{R}^{C \times C}}
$}
\end{equation}
where $\varepsilon$ is the noise rate which controls the ratio of noisy labels.
Without loss of generality, the accuracy (ACC) defined as
\begin{equation}\label{ERRR3}
{\rm{ACC}} =\frac{1}{{N}}\sum\limits_{i = 1}^{{N}} {{{ \delta \left( { y_{i}, \hat y_{i}} \right)}}}  ,
\end{equation}
is adopted to measure the performance of different methods. Herein, $y_{i}$ and $\hat y_{i}$ are respectively the real label and the corresponding prediction of the $i$th sample, and $\delta(x,y)$ is an indicator function with its value being $1$ in the case of $x = y$ and $0$ otherwise.

\subsubsection{Small Scale Classification Data Sets}
First, ten small scale classification data sets are adopted for performance evaluation. The details of these data sets can be found in Table~ \ref{ClassificationData}. In the experiment, the training and testing samples
for each data set are randomly selected and all attributes of data have been
normalized into the range of $[-1,1]$. Meanwhile, the labels of data have been transformed to one-hot vectors. To test the robustness of different
methods, $30\%$ of training data are contaminated by label noise described in \eqref{QQQ}.
\begin{table}[htbp]
	\renewcommand\arraystretch{1.1}
\newcommand{\tabincell}[2]{\begin{tabular}{@{}#1@{}}#2\end{tabular}}
	\caption{Specification of Small Scale Classification Data Sets}
	\begin{center}
		\begin{tabular}{lccccccc}
			\toprule
			Data Set  & Class & Attributes  &  Training  &  Testing\\
			\midrule
			Iris        & 3 & 4 & 75 & 75     \\
			Sonar       & 2 & 60 & 104 & 104  \\
			Smooth Subspace   & 3 & 15 & 150 & 150 \\
			PowerCons	& 2 & 144 & 180 & 180 \\
			Chinatown   & 2 & 24 & 183 & 182  \\
			Dermatolog  & 6 & 34 & 183 & 183 \\
			Balance Scale &	3 & 4 & 313 & 312 \\
            Breast Cancer   & 2 & 10 & 342 & 341  \\
			Australian   & 2 & 14 & 345 & 345 \\			
			Vehicle  & 4 & 18 & 423 & 423 \\	
			\bottomrule
		\end{tabular}
		\label{ClassificationData}
	\end{center}
\end{table}

\begin{table*}[htbp]
	\renewcommand\arraystretch{1.1}
    \newcommand{\tabincell}[2]{\begin{tabular}{@{}#1@{}}#2\end{tabular}}
	\caption{Test ACC of Different Methods on Small Scale Classification Data Sets Averaged Over $20$ Independent Runs (in \%)}
	\begin{center}
		\begin{tabular}{lcccccccccc}
			\toprule
			 Data Set & MSE & MCC & MMCC & MCC-VC & GMCC & KRSL & KMPE & QMEE & ELN \\
			\midrule
			Iris  & 72.13 & 74.13 &	82.87 &	74.13 &	73.07 &	74.13 &	77.47 &	72.47 &	\textbf{86.07} \\
			Sonar & 62.31 & 66.20 &	66.06 &	66.20 &	63.89 &	61.92 &	65.63 &	65.34 &	\textbf{69.52} \\
			Smooth Subspace & 71.17 &	80.63 &	80.87 &	80.63 &	67.87 &	\textbf{80.93} &	79.80 &	76.00 &	80.83 \\
			PowerCons	&   81.08 &	81.67 &	82.47 &	81.67 &	92.89 &	82.69 &	87.83 &	78.17 &	\textbf{94.53} \\
			Chinatown	&   94.18 &	94.12 &	92.36 &	94.12 &	94.40 &	94.23 &	\textbf{95.52} &	92.64 &	95.03 \\
			Dermatolog	&    84.64 &	88.63 &	90.44 &	88.63 &	82.98 &	84.64 &	89.67 &	87.35 &	\textbf{93.52} \\
			Balance Scale &	85.71 &	84.54 &	85.08 &	84.54 &	86.07 &	84.58 &	84.66 &	37.84 &	\textbf{88.06} \\
			Breast Cancer &	93.74 &	94.33 &	94.22 &	94.33 &	93.94 &	94.69 &	93.70 &	93.21 &	\textbf{94.74} \\
			Australian &	83.91 &	84.04 &	85.94 &	84.04 &	82.58 &	85.23 &	84.01 &	74.41 &	\textbf{88.28} \\
			Vehicle	& 69.09 &	69.14 &	69.33 &	69.18 &	69.76 &	\textbf{71.69} &	69.18 &	69.98 &	71.57 	\\
			\bottomrule
		\end{tabular}
		\label{TabClassification}
	\end{center}
\end{table*}
To establish the connection between the attributes and the labels of these data sets, the RBF neural network is adopted as an LIP model for training. Following the idea of RBF, the ${\textbf{h}_i}$ shown in \eqref{E12} can be constructed by
\begin{equation}\label{RBFNN1}
 {\textbf{h}_i} = \left[ {h_1}({\textbf{x}_i}),{h_2}({\textbf{x}_i}), \cdots ,{h_N}({\textbf{x}_i}) \right],
\end{equation}
with
\begin{equation}\label{RBFNN2}
	{h_k}({\textbf{x}_i}) = \kappa_{\sigma}(\textbf{x}_{i},\textbf{x}_{k}), k=1,2,\dots,N,
\end{equation}
where $N$ is the number of training samples, $\kappa_{\sigma}$ is a kernel function controlled by the kernel size $\sigma$. In the following experiments, the Gaussian kernel will be a default choice for the LIP model, and the kernel size $\sigma$ is chosen with MSE criterion from the range of $\{2^{-5},2^{-3},2^{-1},2^{0}, 2^{1}, 2^{3}, 2^{5}\}$.

Instead of always training the ELN with PDF matching method, WOA has been adopted in this part. In addition, we construct the ELN with two extended Gaussian kernels, and hence it can be expressed as
\begin{equation}\label{Experiment1}
L \! =\! \frac{1}{N}\sum\limits_{i\! =\! 1}^N \big(\theta _1{\phi _1} + (1-\theta _1) {\phi _2}\big),
\end{equation}
where $\theta_1$ is the combination coefficient, and $\phi _1$ and $\phi _2$ are constructed by
\begin{equation}\label{Experiment2}
\left\{ \begin{array}{l}
\phi _1 = -\text{exp}(-\lambda_G \mid e_i\mid^{\alpha_{_{G}}})\\
\phi _2 =\frac{1}{\lambda_K}\text{exp} \left( {\lambda_K \left( {1 - {G_{\sigma_K} }(e_i)} \right)} \right) .
\end{array} \right.
\end{equation}
To search the best parameters of ELN with WOA, the fitness function is set as the average error recognition rate of five-fold cross validation on the training set. In addition, the search ranges for $\lambda_G$, $\alpha_{_{G}}$, $\lambda_K$, $\sigma_K$, and $\theta_1$ are set as $[0.01, 2]$, $[1, 3]$, $[0.1, 5]$, $[0.1, 3]$, and $[0, 1]$, respectively.

Table~\ref{TabClassification} shows the test ACC of different methods on the ten classification benchmark data sets\footnote{We perform the classification  process following \cite{RFFEXKRLS}.}. It can be seen from Table~\ref{TabClassification} that, except the data sets ``Smooth Subspace", ``Chinatown" and ``Vehicle``, the RBF network under ELN can always obtain a higher test ACC in comparison with others. Hence, the RBF network developed with ELN can be a good candidate to the classification problems of small scale data sets.

\subsubsection{Large Scale Classification Data Sets}
Then, two large scale data sets, including Mixed National
Institute of Standards and Technology (MNIST) \cite{MNIST} and Fashion-MNIST \cite{MNIST1}, are further adopted to
 test the performance of the proposed method. Both of them include 10 classes,
60 000 training samples, and 10 000 testing samples. Each
sample is represented by an image of 28 $\times$ 28 pixels. The main difference between them is that MNIST is a handwritten digits data set, but Fashion-MNIST involves different categories of clothes. In addition, the recognition of Fashion-MNIST is deemed to be more challenging in comparison with the recognition of MNIST. 

To perform the classification tasks on the two large scale data sets, the label of each image is transformed to be an one-hot vector, and a simple yet effective convolutional neural network model called LeNet5 \cite{MNIST} is chosen for training. Meanwhile, the ELN loss is chosen as the combination of three Gaussian kernels, taking the form of
\begin{equation}\label{Experiment3}
L \! =\! \frac{1}{N}\sum\limits_{i\! =\! 1}^N \big(\theta_1{\phi _1} + \theta_2{\phi _2} + (1-\theta_1 -\theta_2) {\phi_3} \big),
\end{equation}
where $\theta_1$ and $\theta_2$ are combination coefficients; $\phi _1$, $\phi _2$ and $\phi _2$ are set as
\begin{align}\label{Experiment4}
\left\{ \begin{array}{l}
\phi_1 = - \exp ( {  \frac{{{-\parallel \mathbf{e}_i-\mathbf{c}_1\parallel^2}}}{{2{{\sigma} ^2}}}} ) \\
\phi_2 =  -\exp ( {  \frac{{{-\parallel \mathbf{e}_i-\mathbf{c}_2\parallel^2}}}{{2{{\sigma} ^2}}}} ) \\
\phi_3 =  - \exp ( {  \frac{{{-\parallel \mathbf{e}_i-\mathbf{c}_3\parallel^2}}}{{2{{\sigma} ^2}}}} ),
\end{array} \right.
\end{align}
with $\mathbf{c}_1$, $\mathbf{c}_2$ and $\mathbf{c}_3$ denoting the centers of three Gaussian kernels in vector form, and $\sigma$ denotes the kernel size.
In the following experiments, the three centers are simply set to $\mathbf{c}_1 = [0,\cdots,0]\in\mathbb{R}^{1 \times 10}$, $\mathbf{c}_2= [2,\cdots,2]\in\mathbb{R}^{1 \times 10}$, and $\mathbf{c}_3= [-2,\cdots,-2]\in\mathbb{R}^{1 \times 10}$. The kernel size $\sigma$ is chosen from the candidate set of $\{2^{-1}, 2^{3}\}$, and $\theta_1$ and $\theta_2$ are chosen from the candidate set of $\{0, 0.2, 0.6\}$. Once the training is completed, the predicted label of a given test image is determined by the index position of maximum value of the related network outputs. Moreover, it should be noted that the network parameters of LeNet5 are  trained using an SGD optimizer with the assistance of PyTorch 1.10.1. The learning rate of SGD is set to 0.05 at the initial stage, and is changed to 0.005 and 0.0005 after 60 and 120 epochs, respectively. The momentum factor is set to 0.9, and the weight decay factor is set to 0.0001. Meanwhile, the batch size and the number of epochs are set to 128 and 200, respectively.
\begin{table}[htbp]
	\renewcommand\arraystretch{1.3}
	\newcommand{\tabincell}[2]{\begin{tabular}{@{}#1@{}}#2\end{tabular}}
	\caption{Test ACC of different methods on Large Scale Data Sets (in \%)}
	\begin{center}
		\begin{tabular}{lccc}
			\toprule
			Data Set & CE Loss & Weighted CE Loss & ELN \\
			\midrule
			MNIST ($\varepsilon \!=\! 0\%$) & 99.06 &	 99.17 &	\textbf{99.29} \\
			MNIST ($\varepsilon \!=\! 40\%$) &  86.15 &	 96.07 &	\textbf{96.45} \\
			Fashion-MNIST ($\varepsilon \!=\! 0\%$)  & \textbf{91.07}  &	90.80 &	90.79\\
			Fashion-MNIST ($\varepsilon \!=\! 40\%$)  & 77.90  &	80.30 &	\textbf{85.87}\\
			\bottomrule
		\end{tabular}
		\label{TabMNIST}
	\end{center}
\end{table}

Table~\ref{TabMNIST} shows the test ACC of LeNet5 trained with ELN, in which $\varepsilon \!=\! 0\%$ and $\varepsilon \!=\! 40\%$ denote the ratios of noisy labels in training set. For comparison, the test ACCs of LeNet5 trained with cross entropy (CE) loss and a weighted CE loss (designed following \cite{Ref_new01}) are also provided. It can be seen from Table~\ref{TabMNIST} that, when $\varepsilon \!=\! 0\%$, test ACCS of LeNet5 trained with three losses are very close. However, if $40\%$ of training data are contaminated by label noise, the test ACC of LeNet5 trained with CE loss may degenerate, seriously. On the contrary, the networks trained with weighted CE loss and ELN can still obtain relatively higher test ACC. It is worth noting that, although weighted CE loss shows better robustness in comparison with CE loss, it tends to obtain a smaller test ACC compared to ELN when $40\%$ of noisy labels are introduced. The experimental results suggest that ELN is feasible and effective to train deep neural networks when dealing  with the classification problems of large scale data sets.

\section{Conclusion} \label{Sec5}
This paper proposed a novel loss function model called ELN, whose input is an error sample and output is a loss function corresponding to that error sample. It can be verified that ELN provides a unified framework for a large class of error loss functions, including most of ITL loss functions as special cases. Based on the fact that the activation function, weight parameters and network size of ELN can be predetermined or learned from the error samples, we further proposed a new machine learning paradigm where the learning process is divided into two stages: first, learning a loss function using an ELN; second, using the learned loss function to continue to perform the learning. Experimental results on different regression and classification tasks showed that the method based on ELN not only has better performance to deal with non-Gaussian noises than the traditional MSE criterion, but also is very competitive in comparison with some existing robust loss functions.
In the future, more productive technologies to train the ELN model deserve to be further explored.

\appendices

\section{Calculation of $\textbf{K}$ in \eqref{E10}}
Each element of $\textbf{K}$ is formulated as
\begin{equation}\label{APPP1}
\begin{array}{l}
{{\bf{K}}_{ij}} \!=\! \int {{g_{{\sigma _i}}}({c_i})} \;{g_{{\sigma _j}}}({c_j})de\;\;\\
\;\;\;\;\; \!= \!\int {\frac{1}{{\sqrt {2\pi } {\sigma _i}}}\exp ( - \frac{{{{(\varepsilon  - {c_i})}^2}}}{{2{\sigma _i}^2}})\frac{1}{{\sqrt {2\pi } {\sigma _j}}}\exp ( - \frac{{{{(\varepsilon  - {c_j})}^2}}}{{2{\sigma _j}^2}})\;d\varepsilon } \;\\
\;\;\;\;\; \!=\! \frac{1}{{2\pi {\sigma _i}{\sigma _j}}}\int {\exp \left( { - \frac{{{{(\varepsilon  - {c_i})}^2}}}{{2{\sigma _i}^2}} - \frac{{{{(\varepsilon  - {c_j})}^2}}}{{2{\sigma _j}^2}}} \right)\;d\varepsilon } \;\\
\;\;\;\;\; \!=\! r\exp \left( {\frac{{{b^2}}}{{8a{\sigma _i}^2{\sigma _j}^2}}} \right)\int {\exp \left( { - \frac{{a{{\left( {\varepsilon  - \frac{b}{{2a}}} \right)}^2}}}{{2{\sigma _i}^2{\sigma _j}^2}}} \right)\;d\varepsilon } , \;
\end{array}
\end{equation}
where $r$, $a$, and $b$, are, respectively, calculated by
\begin{equation}\label{APPP2}
	\left\{ \begin{array}{l}
		r = \frac{1}{{2\pi {\sigma _i}{\sigma _j}}}\exp \left( { - \frac{{({\sigma _j}^2{c_i}^2 + {\sigma _i}^2{c_j}^2)}}{{2{\sigma _i}^2{\sigma _j}^2}}} \right)\\
		a = {\sigma _j}^2 + {\sigma _i}^2\\
		b = 2({c_i}{\sigma _j}^2 + {c_j}{\sigma _i}^2),\\
	\end{array} \right.
\end{equation}
Let $x = \varepsilon  - \frac{b}{{2a}}$ and ${\sigma ^2} = {\sigma _i}^2{\sigma _j}^2/a$. Then, \eqref{APPP1} can be rewritten as 
\begin{equation}\label{APPP3}
\begin{array}{l}
\;{{\bf{K}}_{ij}} = r\exp \left( {\frac{{{b^2}}}{{8a{\sigma _i}^2{\sigma _j}^2}}} \right)\int {\exp \left( { - \frac{{{x^2}}}{{2({\sigma _i}^2{\sigma _j}^2/a)}}} \right)\;dx} \;\;\\
\;\;\;\;\;\; = r\exp \left( {\frac{{{b^2}}}{{8a{\sigma _i}^2{\sigma _j}^2}}} \right)\int {\exp \left( { - \frac{{{x^2}}}{{2{\sigma ^2}}}} \right)\;dx} \;\;\;\\
\;\;\;\;\;\; = r\exp \left( {\frac{{{b^2}}}{{8a{\sigma _i}^2{\sigma _j}^2}}} \right)\sqrt {2\pi } \sigma \\
\;\;\;\;\;\; = \frac{1}{{\sqrt {2\pi ({\sigma _j}^2 + {\sigma _i}^2)} }}\exp \left( {\frac{{ - \left( { - 2{c_i}{c_j} + {c_i}^2 + {c_j}^2} \right)}}{{2({\sigma _j}^2 + {\sigma _i}^2)}}} \right)\\
\;\;\;\;\;\;  = {G_{\sqrt {{\sigma _j}^2 + {\sigma _i}^2} }}\left( {{c_i} - {c_j}} \right).
\end{array}
\end{equation}

\ifCLASSOPTIONcaptionsoff
  \newpage
\fi



\bibliographystyle{IEEEtran}
\bibliography{References_ELN}

\begin{thebibliography}{10}
\providecommand{\url}[1]{#1}
\csname url@samestyle\endcsname
\providecommand{\newblock}{\relax}
\providecommand{\bibinfo}[2]{#2}
\providecommand{\BIBentrySTDinterwordspacing}{\spaceskip=0pt\relax}
\providecommand{\BIBentryALTinterwordstretchfactor}{4}
\providecommand{\BIBentryALTinterwordspacing}{\spaceskip=\fontdimen2\font plus
\BIBentryALTinterwordstretchfactor\fontdimen3\font minus
  \fontdimen4\font\relax}
\providecommand{\BIBforeignlanguage}[2]{{%
\expandafter\ifx\csname l@#1\endcsname\relax
\typeout{** WARNING: IEEEtran.bst: No hyphenation pattern has been}%
\typeout{** loaded for the language `#1'. Using the pattern for}%
\typeout{** the default language instead.}%
\else
\language=\csname l@#1\endcsname
\fi
#2}}
\providecommand{\BIBdecl}{\relax}
\BIBdecl

\bibitem{MAE}
E.~J. {Coyle} and J.~H. {Lin}, ``Stack filters and the mean absolute error
  criterion,'' \emph{IEEE Trans. Acoust. Speech, Signal Process.}, vol.~36,
  no.~8, pp. 1244--1254, Aug. 1988.

\bibitem{MPE1}
{S. C. Pei} and {C. C. Tseng}, ``Least mean p-power error criterion for
  adaptive \textnormal{FIR} filter,'' \emph{IEEE J. Sel. Areas Commun.},
  vol.~12, no.~9, pp. 1540--1547, Dec. 1994.

\bibitem{MPE2}
J.~Yang, F.~Ye, H.-J. Rong, and B.~Chen, ``Recursive least mean p-power extreme
  learning machine,'' \emph{Neural Netw.}, vol.~91, pp. 22--33, Mar. 2017.

\bibitem{MPE3}
F.~Wen, ``Diffusion least-mean p-power algorithms for distributed estimation in
  alpha-stable noise environments,'' \emph{Electron. Lett.}, vol.~49, no.~21,
  pp. 1355--1356, Oct. 2013.

\bibitem{Huber1}
C.~Chen, C.~Yan, N.~Zhao, B.~Guo, and G.~Liu, ``A robust algorithm of support
  vector regression with a trimmed \textnormal{Huber} loss function in the
  primal,'' \emph{Soft Comput.}, vol.~21, no.~18, pp. 5235--5243, Jun. 2017.

\bibitem{Huber2}
A.~{Esmaeili} and F.~{Marvasti}, ``A novel approach to quantized matrix
  completion using \textnormal{Huber} loss measure,'' \emph{IEEE Signal
  Process. Lett.}, vol.~26, no.~2, pp. 337--341, Feb. 2019.

\bibitem{Logarithmic1}
M.~O. {Sayin}, N.~D. {Vanli}, and S.~S. {Kozat}, ``A novel family of adaptive
  filtering algorithms based on the logarithmic cost,'' \emph{IEEE Trans.
  Signal Process.}, vol.~62, no.~17, pp. 4411--4424, Jun. 2014.

\bibitem{Logarithmic2}
T.~A. {Courtade} and T.~{Weissman}, ``Multiterminal source coding under
  logarithmic loss,'' \emph{IEEE Trans. Inf. Theory}, vol.~60, no.~1, pp.
  740--761, Nov. 2014.

\bibitem{Logarithmic3}
C.~Xu, D.~Tao, and C.~Xu, ``Multi-view intact space learning,'' \emph{IEEE
  Trans. Pattern Anal. Mach. Intell.}, vol.~37, no.~12, pp. 2531--2544, 2015.

\bibitem{ITL}
J.~C. Pr\'{\i}ncipe, \emph{Information Theoretic Learning: \textnormal{Renyi}'s
  Entropy and Kernel Perspectives}.\hskip 1em plus 0.5em minus 0.4em\relax New
  York, NY, USA: Springer, 2010.

\bibitem{MEE-Supervised}
D.~{Erdogmus} and J.~C. {Principe}, ``An error-entropy minimization algorithm
  for supervised training of nonlinear adaptive systems,'' \emph{IEEE Trans.
  Signal Process.}, vol.~50, no.~7, pp. 1780--1786, Jul. 2002.

\bibitem{MEE-ADALINE}
B.~Chen, Y.~Zhu, and J.~Hu, ``Mean-square convergence analysis of
  \textnormal{ADALINE} training with minimum error entropy criterion,''
  \emph{IEEE Trans. Neural Netw.}, vol.~21, no.~7, pp. 1168--1179, Jun. 2010.

\bibitem{P-MCC}
W.~Liu, P.~P. Pokharel, and J.~C. Pr\'{\i}ncipe, ``Correntropy: properties and
  applications in non-\textnormal{Gaussian} signal processing,'' \emph{IEEE
  Trans. Signal Process.}, vol.~55, no.~11, pp. 5286--5298, Nov. 2007.

\bibitem{MEE-Robustness}
B.~Chen, L.~Xing, B.~Xu, H.~Zhao, and J.~C. Pr¨ªncipe, ``Insights into the
  robustness of minimum error entropy estimation,'' \emph{IEEE Trans. Neural
  Netw. Learn. Syst.}, vol.~29, no.~3, pp. 731--737, Dec. 2018.

\bibitem{MCC-Robustness1}
B.~Chen, L.~Xing, H.~Zhao, S.~Du, and J.~C. Príncipe, ``Effects of outliers on
  the maximum correntropy estimation: A robustness analysis,'' \emph{IEEE
  Trans. Syst. Man Cybern. -Syst.}, vol.~51, no.~6, pp. 4007--4012, Aug. 2021.

\bibitem{MCC-Robustness2}
L.~Bako, ``Robustness analysis of a maximum correntropy framework for linear
  regression,'' \emph{Automatica}, vol.~87, pp. 218--225, Jun. 2018.

\bibitem{QMEE}
B.~Chen, L.~Xing, N.~Zheng, and J.~C. Pr\'{\i}ncipe, ``Quantized minimum error
  entropy criterion,'' \emph{IEEE Trans. Neural Netw. Learn. Syst.}, vol.~30,
  no.~5, pp. 1370--1380, Sep. 2019.

\bibitem{GMCC}
B.~Chen, L.~Xing, H.~Zhao, N.~Zheng, and J.~C. Pr\'{\i}ncipe, ``Generalized
  correntropy for robust adaptive filtering,'' \emph{IEEE Trans. Signal
  Process.}, vol.~64, no.~13, pp. 3376--3387, Jul. 2016.

\bibitem{MMCC1}
B.~Chen, X.~Wang, N.~Lu, S.~Wang, J.~Cao, and J.~Qin, ``Mixture correntropy for
  robust learning,'' \emph{Pattern Recognit.}, vol.~79, pp. 318--327, Jul.
  2018.

\bibitem{MMCC2}
Y.~Wang, L.~Yang, and Q.~Ren, ``A robust classification framework with mixture
  correntropy,'' \emph{Inf. Sci.}, vol. 491, pp. 306--318, Jul. 2019.

\bibitem{KRSL}
B.~{Chen}, L.~{Xing}, B.~{Xu}, H.~{Zhao}, N.~{Zheng}, and J.~C. {Principe},
  ``Kernel risk-sensitive loss: Definition, properties and application to
  robust adaptive filtering,'' \emph{IEEE Trans. Signal Process.}, vol.~65,
  no.~11, pp. 2888--2901, Feb. 2017.

\bibitem{KMPE}
B.~{Chen}, L.~{Xing}, X.~{Wang}, J.~{Qin}, and N.~{Zheng}, ``Robust learning
  with kernel mean $p$-power error loss,'' \emph{IEEE Trans. Cybern.}, vol.~48,
  no.~7, pp. 2101--2113, Jul. 2018.

\bibitem{MCC-VC}
B.~Chen, X.~Wang, Y.~Li, and J.~C. Principe, ``Maximum correntropy criterion
  with variable center,'' \emph{IEEE Signal Process. Lett.}, vol.~26, no.~8,
  pp. 1212--1216, Jun. 2019.

\bibitem{MMKCC}
B.~Chen, Y.~Xie, X.~Wang, Z.~Yuan, P.~Ren, and J.~Qin, ``Multikernel
  correntropy for robust learning,'' \emph{IEEE Trans. Cybern.}, doi:
  10.1109/TCYB.2021.3110732.

\bibitem{MCC-Regression}
Y.~Feng, X.~Huang, L.~Shi, Y.~Yang, and J.~A. Suykens, ``Learning with the
  maximum correntropy criterion induced losses for regression,'' \emph{J. Mach.
  Learn. Res.}, vol.~16, no.~30, pp. 993--1034, May 2015.

\bibitem{KMC}
S.~Zhao, B.~Chen, and J.~C. Pr\'{\i}ncipe, ``Kernel adaptive filtering with
  maximum correntropy criterion,'' in \emph{The 2011 International Joint
  Conference on Neural Networks}, 2011, pp. 2012--2017.

\bibitem{RIF-MCC}
Y.~Wang, W.~Zheng, S.~Sun, and L.~Li, ``Robust information filter based on
  maximum correntropy criterion,'' \emph{J. Guid. Control Dyn.}, vol.~39,
  no.~5, pp. 1126--1131, Jan. 2016.

\bibitem{MCUF}
X.~Liu, B.~Chen, B.~Xu, Z.~Wu, and P.~Honeine, ``Maximum correntropy unscented
  filter,'' \emph{Int. J. Syst. Sci.}, vol.~48, no.~8, pp. 1607--1615, Jan.
  2017.

\bibitem{KRMC}
Z.~Wu, J.~Shi, X.~Zhang, W.~Ma, and B.~Chen, ``Kernel recursive maximum
  correntropy,'' \emph{Signal Process.}, vol. 117, pp. 11--16, Dec. 2015.

\bibitem{RHAF-MCC}
Z.~Wu, S.~Peng, B.~Chen, and H.~Zhao, ``Robust \textnormal{Hammerstein}
  adaptive filtering under maximum correntropy criterion,'' \emph{Entropy},
  vol.~17, no.~10, pp. 7149--7166, Oct. 2015.

\bibitem{SMCC}
W.~Ma, H.~Qu, G.~Gui, L.~Xu, J.~Zhao, and B.~Chen, ``Maximum correntropy
  criterion based sparse adaptive filtering algorithms for robust channel
  estimation under non-\textnormal{Gaussian} environments,'' \emph{J. Franklin
  Inst.}, vol. 352, no.~7, pp. 2708 -- 2727, Jul. 2015.

\bibitem{MCKF}
B.~Chen, X.~Liu, H.~Zhao, and J.~C. Pr\'{\i}ncipe, ``Maximum correntropy
  \textnormal{Kalman} filter,'' \emph{Automatica}, vol.~76, pp. 70 -- 77, Feb.
  2017.

\bibitem{MCC-PCA}
R.~{He}, B.~{Hu}, W.~{Zheng}, and X.~{Kong}, ``Robust principal component
  analysis based on maximum correntropy criterion,'' \emph{IEEE Trans. Image
  Process.}, vol.~20, no.~6, pp. 1485--1494, Jun. 2011.

\bibitem{MCC-2D}
G.~{Xu}, S.~{Du}, and J.~{Xue}, ``Precise \textnormal{2D} point set
  registration using iterative closest algorithm and correntropy,'' in
  \emph{2016 International Joint Conference on Neural Networks (IJCNN)}, 2016,
  pp. 4627--4631.

\bibitem{KMEE}
B.~Chen, Z.~Yuan, N.~Zheng, and J.~C. Pr\'{\i}ncipe, ``Kernel minimum error
  entropy algorithm,'' \emph{Neurocomputing}, vol. 121, pp. 160 -- 169, Dec.
  2013.

\bibitem{QMEE-Granger}
B.~{Chen}, R.~{Ma}, S.~{Yu}, S.~{Du}, and J.~{Qin}, ``\textnormal{Granger}
  causality analysis based on quantized minimum error entropy criterion,''
  \emph{IEEE Signal Process. Lett.}, vol.~26, no.~2, pp. 347--351, Jun. 2019.

\bibitem{MCC-Granger}
I.~{Park} and J.~C. {Principe}, ``Correntropy based \textnormal{Granger}
  causality,'' in \emph{2008 IEEE International Conference on Acoustics, Speech
  and Signal Processing}, 2008, pp. 3605--3608.

\bibitem{Parzen1962}
E.~Parzen, ``On estimation of a probability density function and mode,''
  \emph{Ann. Math. Statist.}, vol.~33, no.~3, pp. 1065--1076, 09 1962.

\bibitem{RMEEL}
Y.~Li, B.~Chen, N.~Yoshimura, and Y.~Koike, ``Restricted minimum error entropy
  criterion for robust classification,'' \emph{IEEE Trans. Neural Netw. Learn.
  Syst.}, doi: 10.1109/TNNLS.2021.3082571.

\bibitem{PSO2008}
Y.~del Valle, G.~K. Venayagamoorthy, S.~Mohagheghi, J.~C. Hernandez, and R.~G.
  Harley, ``Particle swarm optimization: Basic concepts, variants and
  applications in power systems,'' \emph{IEEE Trans. Evol. Comput.}, vol.~12,
  no.~2, pp. 171--195, Mar. 2008.

\bibitem{WOA}
S.~Mirjalili and A.~Lewis, ``The whale optimization algorithm,'' \emph{Adv.
  Eng. Softw.}, vol.~95, pp. 51--67, May 2016.

\bibitem{CCQKLMS}
Y.~{Zheng}, S.~{Wang}, Y.~{Feng}, W.~{Zhang}, and Q.~{Yang}, ``Convex
  combination of quantized kernel least mean square algorithm,'' in \emph{2015
  Sixth International Conference on Intelligent Control and Information
  Processing (ICICIP)}, 2015, pp. 186--190.

\bibitem{RBF1988}
D.~Broomhead and D.~Lowe, ``Multivariable functional interpolation and adaptive
  networks,'' \emph{Complex Systems}, vol.~2, pp. 3191--3215, 1988.

\bibitem{RBF1989}
D.~{Lowe}, ``Adaptive radial basis function nonlinearities, and the problem of
  generalisation,'' in \emph{First IEE International Conference on Artificial
  Neural Networks}, 1989, pp. 171--175.

\bibitem{RS}
W.~Deng, Y.~S. Ong, and Q.~Zheng, ``A fast reduced kernel extreme learning
  machine,'' \emph{Neural. Netw.}, vol.~76, pp. 29--38, Apr. 2016.

\bibitem{K-means}
M.~Singhal and S.~Shukla, ``Centroid selection in kernel extreme learning
  machine using k-means,'' in \emph{5th International Conference on Signal
  Processing and Integrated Networks (SPIN)}, 2018, pp. 708--711.

\bibitem{PRQ}
Z.~Qin, B.~Chen, Y.~Gu, N.~Zheng, and J.~C. Principe, ``Probability density
  rank-based quantization for convex universal learning machines,'' \emph{IEEE
  Trans. Neural Netw. Learn. Syst.}, vol.~31, no.~8, pp. 3100--3113, Sep. 2020.

\bibitem{RVFLN1992}
Y.~H. Pao and Y.~Takefuji, ``Functional-link net computing: theory, system
  architecture, and functionalities,'' \emph{Computer}, vol.~25, no.~5, pp.
  76--79, May. 1992.

\bibitem{ELM-K}
G.~Huang, H.~Zhou, X.~Ding, and R.~Zhang, ``Extreme learning machine for
  regression and multiclass classification,'' \emph{IEEE Trans. Syst. Man
  Cybern. Part B}, vol.~42, no.~2, pp. 513--529, Apr. 2012.

\bibitem{BLS}
C.~L.~P. Chen and Z.~Liu, ``Broad learning system: An effective and efficient
  incremental learning system without the need for deep architecture,''
  \emph{IEEE Trans. Neural Netw. Learn. Syst.}, vol.~29, no.~1, pp. 10--24,
  Jul. 2018.

\bibitem{FP-Theory}
R.~P. Agarwal, M.~Meehan, and D.~O'Regan, \emph{Fixed Point Theory and
  Applications}.\hskip 1em plus 0.5em minus 0.4em\relax Cambridge, U.K.:
  Cambridge University Press, 2001.

\bibitem{FP-MCCT}
B.~Chen, J.~Wang, H.~Zhao, N.~Zheng, and J.~C. Principe, ``Convergence of a
  fixed-point algorithm under maximum correntropy criterion,'' \emph{IEEE
  Signal Process. Lett.}, vol.~22, no.~10, pp. 1723--1727, May 2015.

\bibitem{FP-MCC-MEE}
A.~R. Heravi and G.~A. Hodtani, ``A new robust fixed-point algorithm and its
  convergence analysis,'' \emph{J. Fixed Point Theory Appl.}, vol.~19, no.~4,
  pp. 3191--3215, month=, 2017.

\bibitem{RFFEXKRLS}
Z.~Xi, J.~Yang, Y.~Zheng, H.~Wu, and B.~Chen, ``Random fourier features based
  extended kernel recursive least squares with application to \textnormal{fMRI}
  decoding,'' in \emph{2018 Chinese Automation Congress (CAC)}, 2018, pp.
  1390--1395.

\bibitem{UCI2017}
\BIBentryALTinterwordspacing
D.~Dua and C.~Graff, ``\textnormal{UCI} machine learning repository,'' 2017.
  [Online]. Available: \url{http://archive.ics.uci.edu/ml}
\BIBentrySTDinterwordspacing

\bibitem{Ref_new02}
B.~Han, Q.~Yao, X.~Yu, G.~Niu, M.~Xu, W.~Hu, I.~Tsang, and M.~Sugiyama,
  ``Co-teaching: Robust training of deep neural networks with extremely noisy
  labels,'' in \emph{32nd Conference on Neural Information Processing Systems
  (NeurIPS 2018)}, 2018, pp. 8535--8545.

\bibitem{MNIST}
Y.~Lecun, L.~Bottou, Y.~Bengio, and P.~Haffner, ``Gradient-based learning
  applied to document recognition,'' \emph{Proceedings of the IEEE}, vol.~86,
  no.~11, pp. 2278--2324, Nov. 1998.

\bibitem{MNIST1}
\BIBentryALTinterwordspacing
H.~Xiao, K.~Rasul, and R.~Vollgraf, ``Fashion-mnist: a novel image dataset for
  benchmarking machine learning algorithms,'' 2017. [Online]. Available:
  \url{https://arxiv.org/abs/1708.07747}
\BIBentrySTDinterwordspacing

\bibitem{Ref_new01}
T.~Liu and D.~Tao, ``Classification with noisy labels by importance
  reweighting,'' \emph{IEEE Trans. Pattern Anal. Mach. Intell.}, vol.~38,
  no.~3, pp. 447--461, Jul. 2016.

\end{thebibliography}
\end{document}